\documentclass[fleqn,10pt]{wlscirep}
\usepackage[utf8]{inputenc}
\usepackage[T1]{fontenc}
\usepackage{caption}
\usepackage{geometry}
\usepackage{float}
\usepackage{mathtools}
\usepackage{empheq}

\usepackage{lmodern}
\usepackage{array}
\usepackage{makecell}

\usepackage{caption}
\usepackage{subcaption}
\usepackage{graphicx}

\usepackage{array,booktabs}
\newcolumntype{M}[1]{>{\centering\arraybackslash}m{#1}}
\usepackage{slashbox}
\usepackage{colortbl}
\usepackage{etoc}
\usepackage{datetime}

\title{Inductive detection of Influence Operations via Graph Learning}

\author[1]{Nicholas A. Gabriel}
\author[2]{David A. Broniatowski}
\author[1]{Neil F. Johnson}
\affil[1]{Department of Physics, The George Washington University, Washington, DC 20052, USA}
\affil[2]{Department of Engineering Management and Systems Engineering, The George Washington University, Washington, DC 20052, USA}

\date{May 25, 2023}

\begin{abstract}
Influence operations are large-scale efforts to manipulate public opinion. The rapid detection and disruption of these operations is critical for healthy public discourse. Emergent AI technologies may enable novel operations which evade current detection methods and influence public discourse on social media with greater scale, reach, and specificity. New methods with inductive learning capacity will be needed to identify these novel operations before they indelibly alter public opinion and events. We develop an inductive learning framework which: 1) determines content- and graph-based indicators that are not specific to any operation; 2) uses graph learning to encode abstract signatures of coordinated manipulation; and 3) evaluates generalization capacity by training and testing models across operations originating from Russia, China, and Iran. We find that this framework enables strong cross-operation generalization while also revealing salient indicators—illustrating a generic approach which directly complements transductive methodologies, thereby enhancing detection coverage.

\end{abstract}


\begin{document}

\flushbottom
\maketitle
%
%
\thispagestyle{empty}

\section{Introduction}

Manipulation of public opinion by state-backed entities is an ongoing concern. Several influence operations (IO) campaigns intended to shape geopolitical discourse have been identified on various platforms—and particularly on social media\cite{Broniatowski2018,zannettou2019let,zhou2019elites,linvill2020troll,rossetti2023bots,FB-CIB-2022-China-Russia,FB-Q2-2022,FB-Q3-2022,FB-Q4-2022,FB-Q1-2023,Twitter-June-2020,Twitter-Feb-2021}. For example, IO campaigns designed to promote fake news, advance nationalistic narratives, and exacerbate political tensions have been detected across social media platforms including Twitter\cite{Broniatowski2018,rossetti2023bots,Twitter-June-2020,Twitter-Feb-2021}, Facebook \cite{etudo2019facebook,FB-CIB-2022-China-Russia,FB-Q2-2022,FB-Q3-2022,FB-Q4-2022,FB-Q1-2023}, Reddit \cite{reddit2019bots,zannettou2019let}, and Gab \cite{zhou2019elites,zannettou2020characterizing}, among others. Identifying and disrupting such campaigns is an ongoing challenge, in large part because positive attribution of foreign influence is time consuming and does not easily scale within or across platforms. Additionally, the rapid development and adoption of generative AI may enable IO to automate behaviours previously achievable only by human actors, disguising activity and enabling novel strategies which have greater efficacy, scale, reach, and specificity. Most methods of detecting IO to this point have relied on identifying and indexing specific indicators of previous campaigns, making these methods inherently transductive. While such methods will continue to play an important role in detecting and constraining IO activity, identifying increasingly novel and sophisticated IO campaigns will require inductive methods which can generalize from previous observations. We present an inductive learning framework, depicted in Figure \ref{fig:flow}, that addresses this challenge by combining data censorship, graph learning, and feature attribution to identify models and indicators that can generalize across operations and across time.

Previous work in detecting influence operations using machine learning has successfully identified a variety of IO campaigns and activity. Broadly speaking, there have been two main approaches: \textit{content-based} and \textit{graph-based}. Some examples of content-based approaches include: Smith et al.\cite{MIT-LL}, who used narratives derived from topic models to classify Twitter IO accounts in French and English speaking networks; and Alizadeh et al.\cite{Alizadeh2020} who used post text and URL information to classify Twitter posts as belonging to IO or not. Examples of graph-based approaches include: Monti et al.\cite{TwitterGDL}, who used graph networks (GNs) to classify URLs as fake news or not; Vargas et al.\cite{Vargas2021}, who used graph data to classify IO accounts on Twitter which display coordinated behaviour; and Smith et al. \cite{MIT-LL}, who used a network discovery algorithm followed by causal impact estimation to understand the role of individual accounts in propagating IO narratives. 

A previously distinct line of research in cybersecurity, kill chain analysis\cite{KillChainOO,KillChainU,AI_disinfo}, focuses on identifying and disrupting threat actors at each phase of their operation. This approach formalizes various sequences of tactics, techniques, and procedures (TTPs) which IO and other cybercrime operations use to achieve their objectives. In particular, online operations kill chains enable the development of technical indicators which are signatures of cybercrime operations at various phases. These indicators can be used to detect future operations, identify abstract themes across campaigns, analyze trends, and compare TTPs across different operations and time periods. 

These lines of research, as well as reports directly from social media companies, have elucidated a wide range of IO targets, objectives, strategies, and tactics. Many tactics involve the spread of malicious URLs\cite{FB-Q2-2022}, state-backed media, mis/disinformation\cite{GraphikaBadRep}, and particular narratives (e.g., pro-Russian narratives surrounding the Ukrainian war\cite{FB-CIB-2022-China-Russia,FB-Q3-2022,FB-Q4-2022}); other tactics include near-simultaneous link sharing\cite{Giglietto2020}, troll farming\cite{FB-Q2-2022}, mass promotion of particular narratives\cite{MIT-LL,FB-CIB-2022-China-Russia,FB-Q2-2022}, mass reporting of accounts and content\cite{FB-Q2-2022,FB-Q4-2022}, and mass spamming or ``brigading" of specific pages, posts, and users\cite{FB-Q2-2022}. Identifying these tactics has enabled well-resourced social media companies such as Twitter, Meta, and Google to automate the detection of new campaigns that reuse TTPs on their respective platforms. This automation has in turn enabled rapid detection and response to coordinated IO activity. 

Automated detection has greatly constrained the preferred tactics available to IO on relatively well-regulated platforms such as Facebook and Twitter. For example, networks of coordinated and near-simultaneous link sharing (<1 min. apart) are now quickly and routinely removed from these platforms\cite{FB-Q1-2023,FB-Q4-2022,FB-Q3-2022,FB-Q2-2022,FB2021,Twitter-June-2020}. However, this conspicuous behaviour persists as an IO tactic due to the fact that social media ranking algorithms up-rank content with higher engagement, with immediate engagement having an outsized effect on relative ranking and ultimate reach of content\cite{uprank2010}. Hence, to artificially amplify specific narratives during critical periods, IO preferentially coordinate on very short timescales, even at risk of being detected. So while near-simultaneous coordination may be largely curtailed by platforms or even abandoned by IO in the future, coordination on short timescales is expected to continue. For particularly sophisticated IO networks, one would expect that future coordination patterns would mimic that of authentic users.

\begin{figure}[H]
\begin{picture}(11,11)
\put(44,104){ \makecell{coordination \\ time}}
\put(20,64){ \makecell{fake account \\ type}}
\put(130,-150){ {\Large \rotatebox{0}{$ \xlongrightarrow[\textrm{decreased efficacy}]{}$}} }
\put(302,20){ {\Large \rotatebox{-90}{$ \xlongrightarrow{\textrm{increased cost}}$}} }
\end{picture}
    \centering
\setlength\tabcolsep{0pt}
\begin{tabular}{|@{\rule[-1.4cm]{0pt}{3.3cm}}*{3}{M{3.3cm}|}} \hline
\backslashbox[1.5cm][l]{  }{}& \textbf{near-simultaneous} (< 1 min.) &  \ \    \textbf{quasi-authentic} \, \ \ \ \   (< 100 min.) \\\hline
\textbf{spam accounts} & \makecell{\textit{immediately} \\ \textit{detectable}} & \makecell{\textit{} \\ \textit{detectable}}  \\ \hline
\textbf{persona building} & \makecell{\textit{} \\ \textit{detectable}} & \makecell{\textit{currently} \\ \textit{undetectable}} \\
    \hline 
    \end{tabular}
    \vspace*{1cm}
\caption{  Current landscape of automated detection on mainstream social platforms. Advances in automated detection will push influence operations towards less effective methods of coordination and more costly approaches to fake account creation. In turn, influence operations may be able to compensate by augmenting existing capabilities with emergent AI systems.}
    \label{fig:squares}
\end{figure}

Fake account detection\cite{Twitter-June-2020,FB-Q2-2022,FB-Q4-2022,FB-Q1-2023} has also greatly improved. In response, IO have tried to obviate detection by crafting realistic profiles that mimic authentic users in a process called \textit{persona building}. The process of persona building has presumably been a manual effort to this point, as smaller numbers of these meticulously crafted fake accounts are observed as part of any IO compared to the much larger numbers of less sophisticated ``spam" accounts (though part of this discrepancy may be a survivorship bias). A common approach to persona building is to mimic existing accounts that promote narratives favorable to the IO objective, such as inflammatory political content promoted by Russian and Iranian IO campaigns leading up to the 2016 and 2020 U.S. presidential elections \cite{Twitter-June-2020,FB2021}. This process of mimicry requires significant investment of human effort, as this approach requires the generation of novel content such as text and images. However, it is not difficult to imagine that in the near future a single IO operative could automate the persona building process using novel AI tools to farm a large number of fake accounts. Indeed the use of GAN produced profile pictures\cite{KillChainOO,FB-Q1-2023} and deep fakes\cite{GraphikaDeepFake} has been reported. While the automated detection of near-simultaneous coordination and fake accounts will push campaigns towards less efficacious and more costly approaches (Figure \ref{fig:squares}), they may be able to compensate with greater scalability, novelty, and specificity enabled by AI.

While mainstream platforms have the resources and desire to improve regulation, alternative platforms are less equipped, and possibly unwilling, to follow suit. Exploiting this situation, the Russian origin Secondary Infektion campaign from 2014-2020 made use of over 300 platforms including WordPress, BlogSpot, Quora, Reddit, and LiveJournal to circulate fake news and seed fabricated primary sources\cite{SecondaryInfektion}. A subsequent Russian campaign from 2020-2022 (likely a continuation of the same operation) targeted 35 alternative platforms that intentionally have little or no regulation such as Gab, Gettr, Parler, and Truth Social\cite{GraphikaBadRep}. While all of these platforms combined have a smaller audience than most mainstream platforms, they demonstrate continued trends of IO in microtargeting specific audiences and diversifying channels of influence. Countering these trends will require methods of detection that can identify operations across platforms, as well as generalize previous observations on mainstream platforms to newly targeted platforms. Additionally, while mainstream platforms have thus far been proactive in identifying and removing inauthentic actors, it is unclear to what extent this will continue to be true. 

Even in light of these trends, continuing to identify and index TTPs for transductive detection will still be paramount to constrain future IO. In other words, the foundation of IO detection will continue to be transductive—or based on specific, previously observed indicators. Transductive approaches will continue to be effective in constraining IO for two main reasons: 1) operations can only develop new TTPs so quickly; 2) previously indexed TTPs often represent the preferred tactics of IO, which they may be slow to abandon. In order to continue shaping public discourse in the near term, one can then expect continued reuse of TTPs, even if these are largely ineffective on mainstream platforms. In the long term, however, one can expect IO to develop novel tactics that avoid detection and reach larger segments of online users. On one hand, this means that future IO will likely have less impact per action (post, like, share, etc.) since they cannot maximally exploit the platforms in which they are embedded. On the other hand, AI systems such as StyleGAN2\cite{stylegan2}, DeepFaceLab\cite{perov2021deepfacelab}, GPT\cite{openai2023gpt4}, and DALL-E\cite{dalle} may allow IO to more easily craft realistic profiles and content, thereby enabling novel campaigns that employ previously costly tactics at greater scale. In such cases, it is unclear how effective transductive methods will be, if at all. Hence, developing inductive methods of detection will be necessary to proactively identify and disrupt novel campaigns which can consequentially alter public opinion in a matter of days (e.g., in the days leading up to an election\cite{Twitter-June-2020,FB2021}). To this end, we observe two fundamental techniques that IO use when manipulating public opinion:
\begin{itemize}
    \item[(I)] linking to off-platform websites that are considered credible by the target audience and/or possessing decreased regulation;
    \item[(II)] coordinated promotion of content supporting specific narratives.
\end{itemize}
Arguably, IO can have little impact on public opinion without employing these techniques in some form. We  use this observation to design (I) content-based and (II) graph-based indicators which are general enough to identify novel campaigns from previous campaigns, and use graph representation learning to encode abstract signatures of coordination from these indicators. In particular, we determine indicators that are not specific to any particular IO campaign by explicitly censoring previously identified content- and graph-based technical indicators. We call indicators resulting from this type of censorship \textit{generalized indicators}, since they will be common across both IO campaigns and authentic users, and also across platforms.


We investigate how specific choices of generalized indicators and graph learning techniques can identify inauthentic actors across IO campaigns, thereby developing a framework that directly complements the transductive methodologies established in previous work. In doing so, we note the correspondence between the generalized indicators used here and previously used technical indicators:

\begin{table}[H]
    \centering
    \begin{tabular}{|c|c|c|}
    \hline
      \textbf{feature type}   & \textbf{technical indicator\cite{KillChainOO,Alizadeh2020,MIT-LL,Vargas2021,Giglietto2020}} & \textbf{generalized indicator}  \\ \hline
       content-based  & \makecell{ political and news domains; \\ URLs containing malware, propaganda, and fake news} &   censored domains \\ \hline
       graph-based & graph size, betweenness, clustering & censored graph learning\\ \hline
       coordination & near-simultaneous  (<1 min.) & quasi-authentic (<100 min.) \\
       \hline
    \end{tabular}
    \caption{Correspondence between previously effective technical indicators and generalized alternatives.}
    \label{tab:generalized}
\end{table}
\noindent Furthermore, we investigate three specific advances of previous approaches: 
\begin{itemize}
    \item[(1)] Identification of content-based and graph-based indicators which enable cross-operation generalization;
    \item[(2)] Utilization of graph learning to encode abstract signatures of coordination, thereby automating graph-based feature engineering and inference;
    \item[(3)] Investigation of a broad coordination window, moving from near-simultaneous (<1 min.) to quasi-authentic (<100 min.) interarrival times.
\end{itemize}
\section{Results}
\begin{figure}[H]
    \centering
    \includegraphics[width=1.\textwidth]{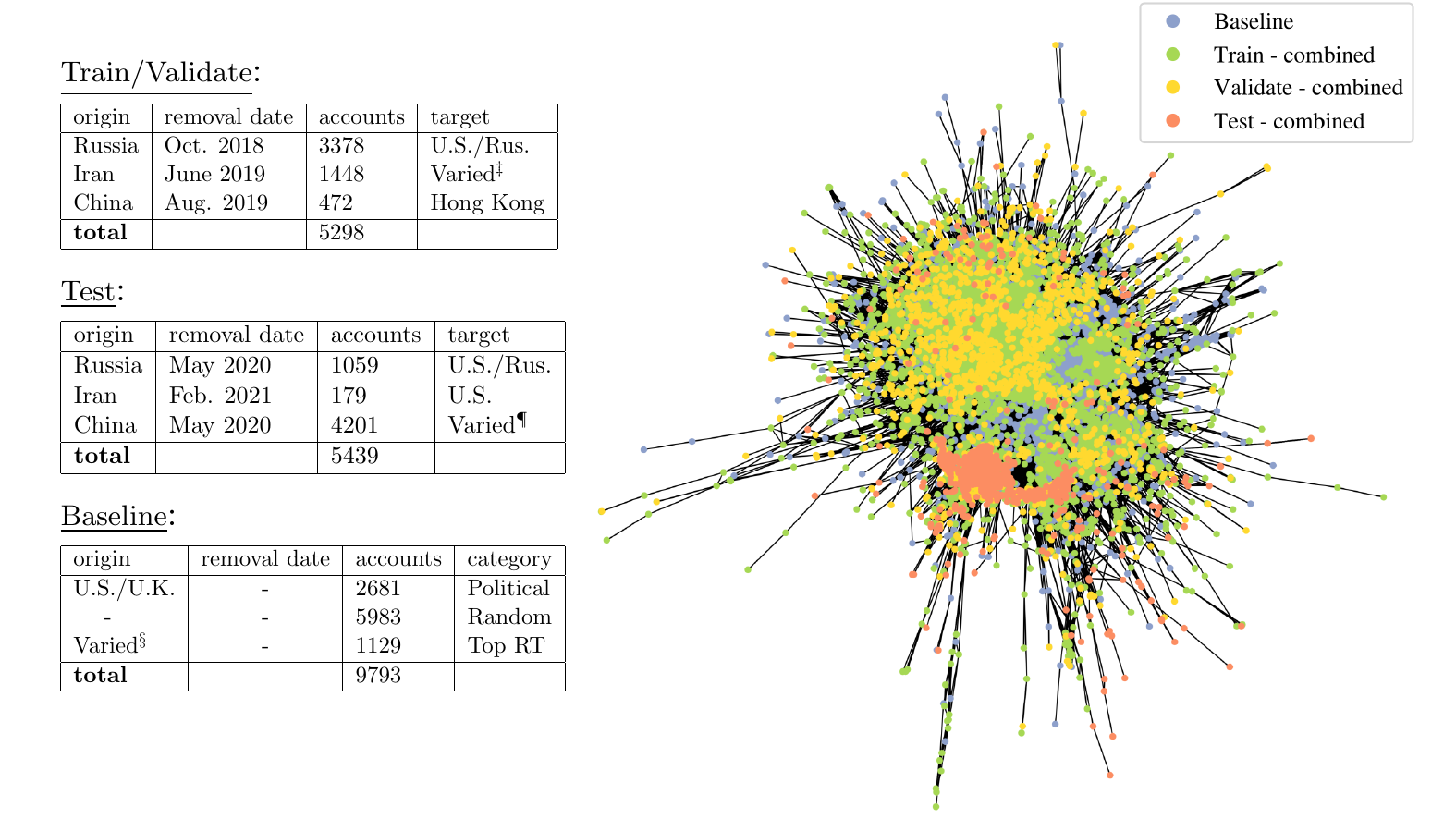}
    \caption{ Composition of training, validation, and test and baseline sets. \\  $ \ \ \ ^\ddag$ target audiences includes the U.S., Latin America, Saudia Arabia, Israel, Indonesia. \\ $ \ \ \ ^{\P}$ target audiences includes the U.S., China, and Russia. \\ $ \ \ \ ^{\S}$ account origins include U.S., Russia, and China }
    \label{fig:net_splits}
\end{figure}

Following the framework presented in Figure \ref{fig:flow}, we assess the extent to which specific machine learning models and generalized indicators can identify IO accounts across campaigns, both intra-operation and inter-operation (results shown in Table \ref{fig:big_table}). For this purpose, we select six IO campaigns (Figure \ref{fig:net_splits}) belonging to three coordinated operations: Russia, China, and Iran; and a comprehensive baseline described in the next section. We analyze intra- and inter-campaign co-URL statistics in Figure \ref{fig:coURL}, demonstrating operation specific trends and the independence of the three operations chosen. In Table \ref{tab:ig_agg}, we determine which indicators enable cross-campaign generalization using an axiomatic attribution method, integrated gradients\cite{IntegratedGradients}.

\subsection{Model and Indicator Evaluation}

We evaluate the effectiveness of several machine learning models—Logistic Regression (LR), Random Forest (RF), Multilayer Perceptron (MLP), Graph Convolutional Network (GCN), deep Message Passing Neural Network (MP-GCN), and shallow Message Passing Neural Network (MP-GCN(s))—on node classification tasks comprising: 10737 influence operation accounts reported by Twitter between 2018 and 2021; and 9793 baseline Twitter accounts not known to be part of any influence operation. The IO accounts were reported in several releases between October 2018 and February 2021, which we split as in Figure \ref{fig:net_splits} to simulate a prediction task on unseen data. The baseline includes accounts which directly impact public discourse (journalists, media outlets, writers, and academics), random accounts, and accounts highly retweeted by the IO training set. Our goal is to differentiate IO accounts versus this baseline using a set of generalized indicators, as well as determine the optimal graph encoding (G.E.) for each model (i.e. which of \textbf{node2vec}, \textbf{Laplacian Eigenmaps}, \textbf{Random Walk Positional Encoding}, and \textbf{Network Features} to include).

In order to assess the generalization capacity of particular model and indicator choices, we formulate two tasks. The first is intra-operation classification (\textbf{A} in Table \ref{fig:big_table}), where we train on a campaign of a particular operation (Russia, China, or Iran) and test on a later identified campaign of same operation. The second task is inter-operation classification (\textbf{B} in Table \ref{fig:big_table}), where we train on all operations \textit{except} the test operation. For the three campaigns in the training/validation set and the test set, this implies three subtasks for \textbf{A} and \textbf{B}. To enable comparison between the two sets of subtasks, we sample the validation and test sets for each respective subtask identically (e.g., the results of task \textbf{A1} and \textbf{B1} can be compared directly). This allows us to assess how well each model can  generalize from independent operations based on any changes in performance from tasks \textbf{A1}, \textbf{A2}, and \textbf{A3} to tasks \textbf{B1}, \textbf{B2}, and \textbf{B3}, respectively.

\begin{figure}[H]
    \centering
    \includegraphics[width=1.05\linewidth]{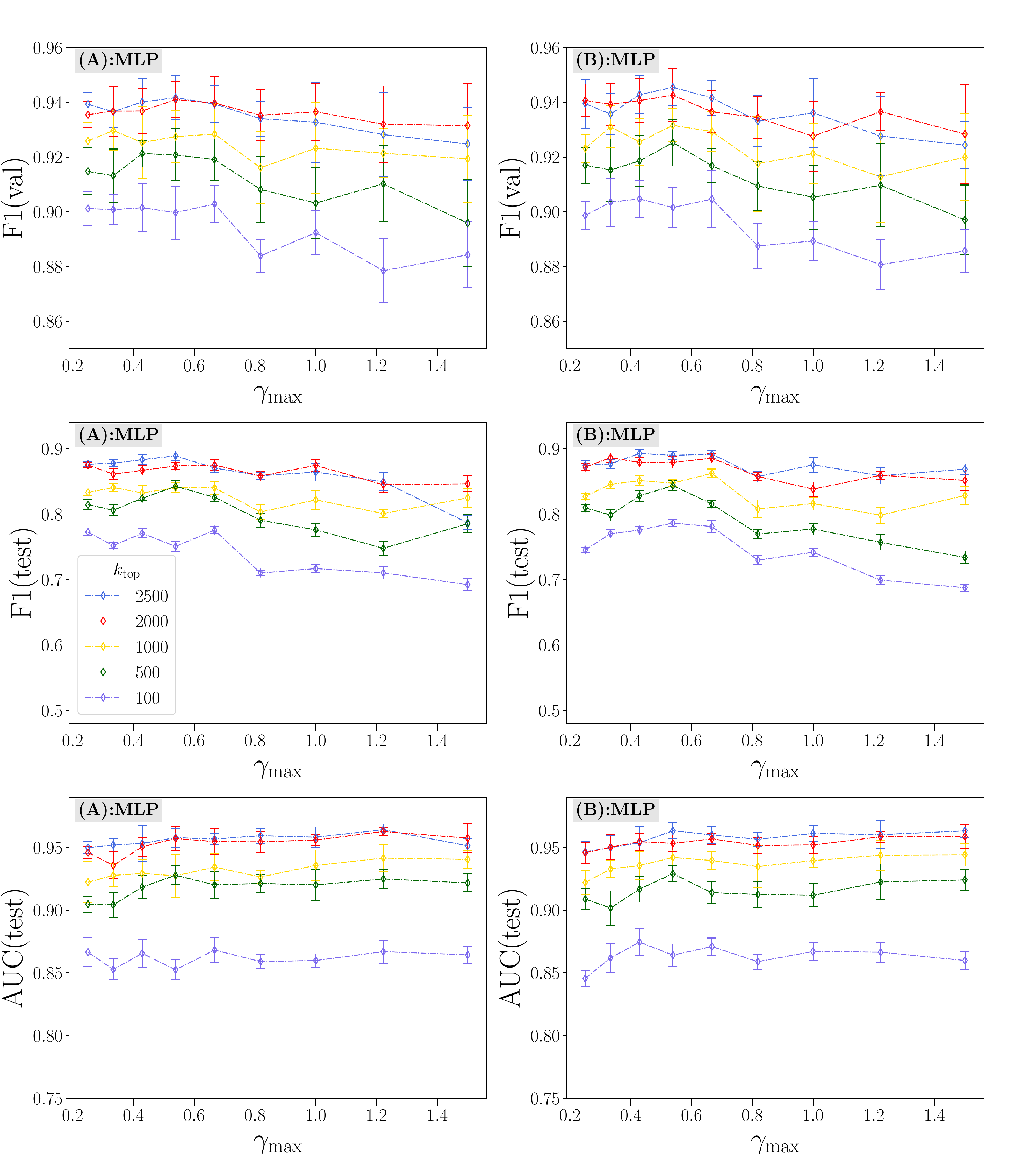}
    \caption{\underline{Top to bottom}: Aggregated F1(val), F1(test), and AUC(test) for inter-operation (\textbf{A}, left column) and inter-operation (\textbf{B}, right column) classification with varying $\gamma_{\textrm{max}}$ and $k_{\textrm{top}}$. Errors are calculated from individual subtasks (\textbf{A1}, \textbf{A2}, \textbf{A3} and \textbf{B1}, \textbf{B2}, \textbf{B3}) using uncertainty propagation.}
    \label{fig:gamma_varied}
\end{figure}

We evaluate the effect of varying the content-based feature set both in terms of stringency ($\gamma_{\textrm{max}}$) and minimum prevalence ($k_{\textrm{top}}$) on model performance in Figure \ref{fig:gamma_varied}. We choose MLP with all graph encodings (G.E. = $\dagger$$\dagger$$\dagger$$\dagger$) as a representative model since it consistently performed well across all subtasks. The effect of increasing $k_{\textrm{top}}$ improves model performance on all metrics in a nearly monotonic manner, ostensibly reaching saturation around $k_{\textrm{top}} = 2000$. The effect of varying the maximum frequency ratio ($\gamma_{\textrm{max}}$) has a more nuanced effect on performance, but a fairly stringent value of $\gamma_{\textrm{max}} \in \{ 0.43, 0.67 \}$ appears to produce greater generalization than smaller or larger values, with increases beyond $\gamma_{\textrm{max}}=0.67$ producing a nearly monotonic decrease in performance for F1(val/test), but not for AUC.

\vspace*{-3.cm}
\newpage

\vspace*{-1.3cm}

\hspace*{-1cm}
\begin{table}[H]
    \begin{minipage}{.52\linewidth}
      \centering
\begin{center}
\begin{tabular}{l}
      \textbf{(A): Combined, intra-operation}   \\  \hline 
\end{tabular}
\end{center}
\vspace*{-10mm}
\begin{center}
\begin{tabular}{l}
\end{tabular}
\end{center}

\begin{tabular}{l|lll|l}
model      & F1(val.)$^{\ddag}$ & F1(test) & AUC(test) & G.E.  \\  \hline 
LR         & 92.92                      & 84.94                      & 85.23 & ${}^{\ast}$$\dagger$$\dagger$$\dagger$  \\
RF         & \underline{\textbf{97.50}} & 86.58                      & 86.90 & ${}^{\ast\ast\ast\ast}$ \\
MLP        & \textbf{95.96}             & \textbf{91.13}             & \underline{\textbf{94.68}} & $\dagger$$\dagger$$\dagger$$\dagger$ \\
GCN & 95.33                      & \underline{\textbf{91.71}} & 91.79 & $\dagger$$\dagger$$\dagger$$\dagger$ \\
MP-GCN(s) & {95.55}                    & {91.02}                    & \textbf{94.44} & $\dagger$$\dagger$$\dagger$$\dagger$  \\
MP-GCN & 95.49                      & 90.64                      & 93.01 & $\dagger$$\dagger$$\dagger$$\dagger$

\end{tabular}
    \end{minipage}%
    \begin{minipage}{.52\linewidth}
      \centering
\begin{center}
\begin{tabular}{l}
      \textbf{(B): Combined, inter-operation}   \\  \hline 
\end{tabular}
\end{center}
\vspace*{-10mm}
\begin{center}
\begin{tabular}{l}
\end{tabular}
\end{center}

\begin{tabular}{l|lll|l}
model      & F1(val.) & F1(test) & AUC(test) & G.E.  \\  \hline 
LR         & 82.97                      & 76.92                      & 77.64  & ${}^{\ast}$$\dagger$$\dagger$$\dagger$  \\
RF         & 82.36                      & 81.06                      & 81.52  & ${}^{\ast\ast\ast\ast}$ \\
MLP        & \textbf{92.04}          & \textbf{89.92}                      & \underline{\textbf{96.59}}  & $\dagger$$\dagger$$\dagger$$\dagger$ \\
GCN & 86.39                      & \underline{\textbf{91.25}} & 93.92  & $\dagger$$\dagger$$\dagger$$\dagger$ \\
MP-GCN(s) & 91.65                      & 88.05                      & \textbf{96.28}  & $\dagger$$\dagger$$\dagger$$\dagger$  \\
MP-GCN & \underline{\textbf{92.40}} & 88.06 & 96.05  & $\dagger$$\dagger$$\dagger$$\dagger$

\end{tabular}
    \end{minipage} 
\end{table}
\vspace*{-.2cm}
\hrule
\vspace*{.05cm}
\hrule
\vspace*{-.1cm}

\begin{table}[H]
\nopagebreak
    \begin{minipage}{.52\linewidth}
    \nopagebreak
      \centering
\begin{center}
\nopagebreak
\begin{tabular}{l}
      \textbf{(A1): \ Rus(18)   $ \ \rightarrow \ $   Rus(18) / Rus(20)}   \\  \hline 
\end{tabular}
\end{center}
\vspace*{-4mm}
\begin{center}
\begin{tabular}{l}
\boxed{N^{(y=1)}_{\textrm{train}} = 2702 \ \ \rightarrow \ \ N^{(y=1)}_{\textrm{val.}/\textrm{test}} = 676/1059}
\end{tabular}
\end{center}

\begin{tabular}{l|lll|l}
model      & F1(val.)$^{\ddag}$ & F1(test) & AUC(test) & G.E.  \\  \hline 
LR         & 96.25   & 90.27   &  90.45 & $\dagger$$\dagger$$\dagger$$\dagger$  \\
RF         & \underline{\textbf{99.09}} & {91.82}     & 92.95  & $\dagger$$\dagger$$\dagger$$\dagger$ \\
MLP        & \textbf{97.27}  & 91.90      & \underline{\textbf{96.72}} & $\dagger$$\dagger$$\dagger$$\dagger$ \\
GCN & 94.80 & 90.10    & 93.83  & $\dagger$$\dagger$$\dagger$$\dagger$ \\
MP-GCN(s) & 96.29  & \textbf{92.80}    & \textbf{95.46} & $\dagger$$\dagger$$\dagger$$\dagger$  \\
MP-GCN & 96.36  & \underline{\textbf{92.84}}    &  95.41 & $\dagger$$\dagger$$\dagger$$\dagger$

\end{tabular}
\centering

\centering
\begin{center}
\begin{tabular}{l}
      \textbf{(A2): \ Chn(19) $ \ \rightarrow \ $ Chn(19) / Chn(20)}   \\  \hline 
\end{tabular}
\end{center}
\vspace*{-4mm}
\begin{center}
\begin{tabular}{l}
\boxed{ N^{(y=1)}_{\textrm{train}} = 377 \ \ \rightarrow \ \ N^{(y=1)}_{\textrm{val.}/\textrm{test}} = 95/4201 }
\end{tabular}
\end{center}

\begin{tabular}{l|lll|l}
model      & F1(val.)$^{\ddag}$ & F1(test)  & AUC(test) & G.E.  \\  \hline 
LR         & 94.54   & 88.08    & 88.30   & ${}^{\ast}$$\dagger$$\dagger$$\dagger$  \\
RF         & \underline{\textbf{97.06}} & 91.75    & 91.87  & ${}^{\ast\ast\ast\ast}$ \\
MLP        & \textbf{95.44}  & 92.63    & \underline{\textbf{93.70}}  & $\dagger$$\dagger$$\dagger$$\dagger$ \\
GCN & 95.40  & 92.83   & 89.26   & $\dagger$$\dagger$$\dagger$$\dagger$ \\
MP-GCN(s) & 95.04  & \underline{\textbf{93.57}}   & \textbf{92.56}  & $\dagger$$\dagger$$\dagger$$\dagger$  \\
MP-GCN & 95.26   & \textbf{93.33}   & 90.15  & $\dagger$$\dagger$$\dagger$$\dagger$
\end{tabular}
\vspace*{0mm}
\centering
\centering
\begin{center}
\begin{tabular}{l}
      \textbf{(A3): \ Iran(19) $ \ \rightarrow \ $ Iran(19) / Iran(21)}   \\  \hline 
\end{tabular}
\end{center}
\vspace*{-4mm}
\begin{center}
\begin{tabular}{l}
\boxed{ N^{(y=1)}_{\textrm{train}} = 1158 \ \ \rightarrow \ \  N^{(y=1)}_{\textrm{val.}/\textrm{test}} = 290/179 }
\end{tabular}
\end{center}

\begin{tabular}{l|lll|l}
model      & F1(val.)$^{\ddag}$ & F1(test)  & AUC(test) & G.E.  \\  \hline 
LR         & 88.36   & 77.60   & 78.01   & ${}^{\ast}$$\dagger$$\dagger$$\dagger$  \\
RF         & \underline{\textbf{96.38}}  & 77.06    & 77.64   & ${}^{\ast\ast\ast\ast}$ \\
MLP        & 95.21  & \textbf{88.95}  & \textbf{93.69}  & $\dagger$$\dagger$$\dagger$$\dagger$ \\
GCN & 95.80 & \underline{\textbf{92.26}}    & 92.40  & $\dagger$$\dagger$$\dagger$$\dagger$ \\
MP-GCN(s) & \textbf{95.33}  & 86.97    & \underline{\textbf{95.35}} & $\dagger$$\dagger$$\dagger$$\dagger$  \\
MP-GCN & 94.87  & 86.13    & 93.61   & $\dagger$$\dagger$$\dagger$$\dagger$
\end{tabular}
    \end{minipage}%
    \begin{minipage}{.52\linewidth}
    \nopagebreak
\begin{center}
\nopagebreak
\begin{tabular}{l}
      \textbf{(B1): \ Chn(19) + Iran(19)  $\rightarrow$ Rus(18) / Rus(20)}   \\  \hline 
\end{tabular}
\end{center}
\vspace*{-4mm}
\begin{center}
\begin{tabular}{l}
\boxed{ N^{(y=1)}_{\textrm{train}} = 1920 \ \ \rightarrow \ \  N^{(y=1)}_{\textrm{val.}/\textrm{test}} = 676/1059 }
\end{tabular}
\end{center}

\begin{tabular}{l|lll|l}
model      & F1(val.) & F1(test)  & AUC(test) & G.E.  \\  \hline 
LR         & \underline{\textbf{91.93}} &  89.91 &   90.11   & ${}^{\ast}$$\dagger$$\dagger$$\dagger$  \\
RF         & 83.50 &  86.46 &    86.71   & ${}^{\ast\ast\ast\ast}$ \\
MLP        & \textbf{90.11} &  90.48 &   \underline{\textbf{96.77}}  & $\dagger$$\dagger$$\dagger$$\dagger$ \\
GCN & 84.64 &  92.22  &  93.63  & $\dagger$$\dagger$$\dagger$$\dagger$ \\
MP-GCN(s) & 86.12 &  \textbf{93.21}  &  94.43 & $\dagger$$\dagger$$\dagger$$\dagger$  \\
MP-GCN & 86.15 &  \underline{\textbf{93.44}}  &  \textbf{94.51}   & $\dagger$$\dagger$$\dagger$$\dagger$
\end{tabular}
\centering

\centering
\begin{center}
\begin{tabular}{l}
      \textbf{(B2): \ Rus(18) + Iran(19)  $\rightarrow$ Chn(19) / Chn(20)}   \\  \hline 
\end{tabular}
\end{center}
\vspace*{-4mm}
\begin{center}
\begin{tabular}{l}
\boxed{ N^{(y=1)}_{\textrm{train}} = 4826 \ \  \rightarrow \ \  N^{(y=1)}_{\textrm{val.}/\textrm{test}} = 95/4201 }
\end{tabular}
\end{center}

\begin{tabular}{l|lll|l}
model      & F1(val.) & F1(test)  & AUC(test) & G.E.  \\  \hline 
LR         & 89.23 &  82.66 &   83.08   & $\dagger$$\dagger$$\dagger$$\dagger$  \\
RF         & 82.31 &  93.47 &   93.54   & ${}^{\ast\ast\ast\ast}$ \\
MLP        & 91.52 &  \textbf{94.63} &   \underline{\textbf{97.82}}  & $\dagger$${}^{\ast\ast}$$\dagger$ \\
GCN & 90.40   & 94.41 &   \textbf{97.51}  & $\dagger$$\dagger$$\dagger$$\dagger$ \\
MP-GCN(s) & \underline{\textbf{92.10}} &  94.16 &   97.38 & $\dagger$$\dagger$$\dagger$$\dagger$  \\
MP-GCN & \textbf{91.79} &  \underline{\textbf{94.78}} &   97.29   & $\dagger$$\dagger$$\dagger$$\dagger$
\end{tabular}
\vspace*{0mm}
\centering
\centering
\begin{center}
\begin{tabular}{l}
      \textbf{(B3): \ Chn(19) + Rus(18)  $\rightarrow$ Iran(19) / Iran(21)}   \\  \hline 
\end{tabular}
\end{center}
\vspace*{-4mm}
\begin{center}
\begin{tabular}{l}
\boxed{ N^{(y=1)}_{\textrm{train}} = 3850 \ \rightarrow \  N^{(y=1)}_{\textrm{val.}/\textrm{test}} = 290/179 }
\end{tabular}
\end{center}

\begin{tabular}{l|lll|l}
model      & F1(val.) & F1(test)  & AUC(test) & G.E.  \\  \hline 
LR         & 73.36 &  71.36 &   72.56   & ${}^{\ast\ast\ast\ast}$  \\
RF         & 71.94  & 75.01  &   75.61   & ${}^{\ast\ast\ast\ast}$ \\
MLP        & \textbf{89.57} &  \textbf{86.58} &   \underline{\textbf{98.41}}  & $\dagger$$\dagger$$\dagger$$\dagger$ \\
GCN & 75.62  & \underline{\textbf{87.39}}  &  92.05  & $\dagger$$\dagger$$\dagger$$\dagger$ \\
MP-GCN(s) & 87.85 &  83.67 &   {96.15} & $\dagger$$\dagger$$\dagger$$\dagger$  \\
MP-GCN & \underline{\textbf{90.65}} &  84.01 &   \textbf{97.37}   & $\dagger$$\dagger$$\dagger$$\dagger$
\end{tabular}
    \end{minipage} 
\caption{\underline{Top}: Aggregated intra-operation (\textbf{A}) and inter-operation (\textbf{B}) results; F1 and ROC-AUC scores are the harmonic mean of the individual subtasks shown in the bottom six tables; G.E. is the median value of subtasks. \underline{Bottom}: Individual subtask results for intra-operation (\textbf{A1}, \textbf{A2}, and \textbf{A3}) and cross-operation (\textbf{B1}, \textbf{B2}, and \textbf{B3}) classification. We note that the validation and test sets in the intra-operation and cross-operation subtasks are sampled identically, and hence can be compared. Each graph encoding (G.E.) denotes the absence (${}^*$) or presence ($\dagger$) of \textbf{node2vec}, \textbf{Laplacian Eigenmaps}, \textbf{Random Walk Positional Encoding}, and \textbf{Network Features} determined from a censored graph. Each model is trained with $\gamma_{\textrm{max}}=0.54$ and $k_{\textrm{top}}=2500$ per Figure \ref{fig:gamma_varied}.   $^{\ddag}$In sample.}
\label{fig:big_table}
\end{table}

\newpage

\subsection{Coordination analysis of composite dataset}
\begin{figure}[H]
\vspace*{-1.5cm}
    \centering
    \begin{subfigure}[t]{0.45\textwidth}
        \centering
        \makebox[\textwidth][c]{\includegraphics[width=1.1\textwidth]{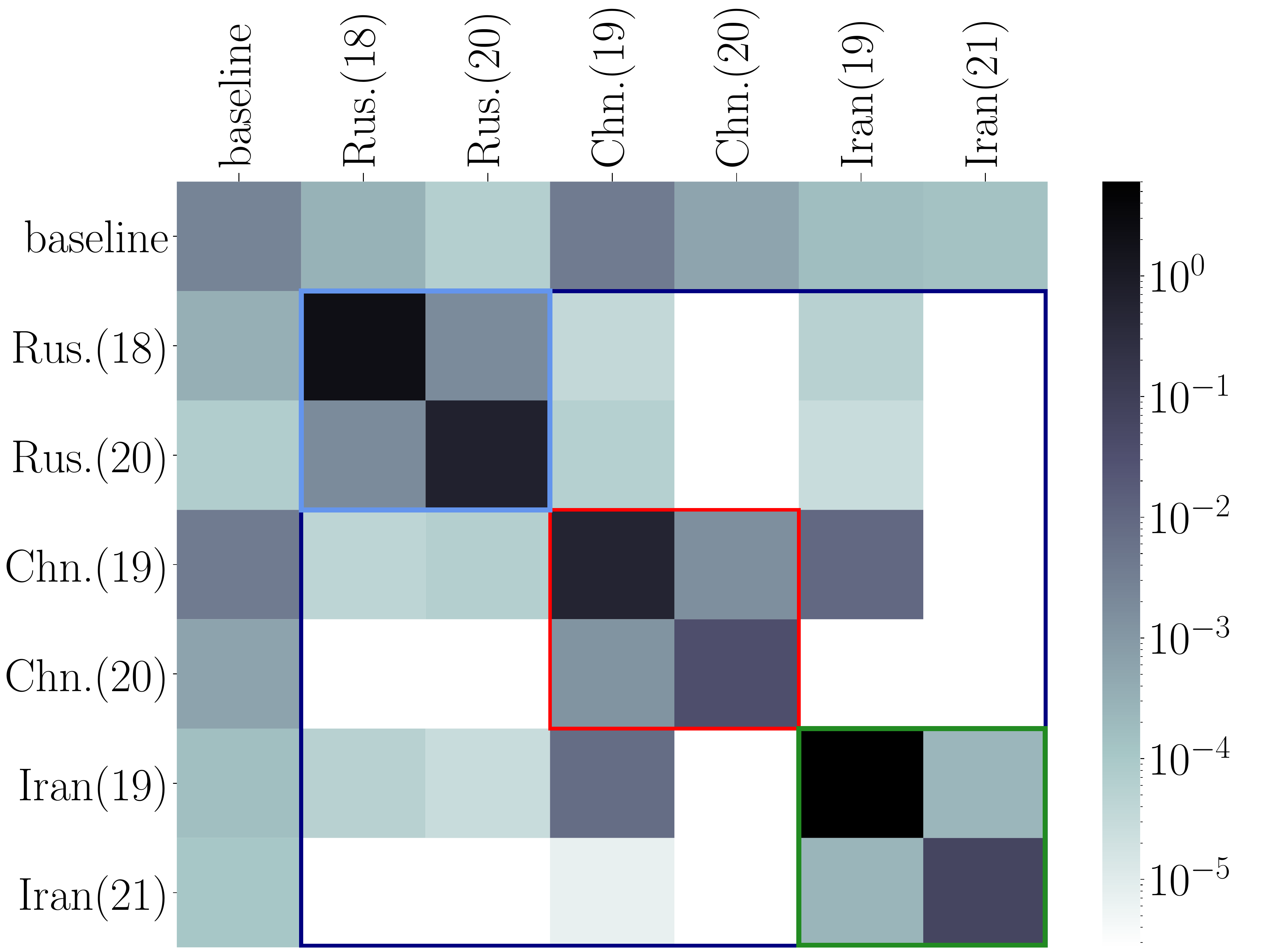}}
        \vspace*{-.3cm}
        \caption{ \centering Frequency of co-URLs between each subset. Counts are normalized by the size of the leading and lagging subsets.} 
    \label{fig:curl}
    \end{subfigure}
    \hfill
    \begin{subfigure}[t]{0.45\textwidth}
        \centering
        \makebox[\textwidth][c]{\includegraphics[width=1.1\textwidth]{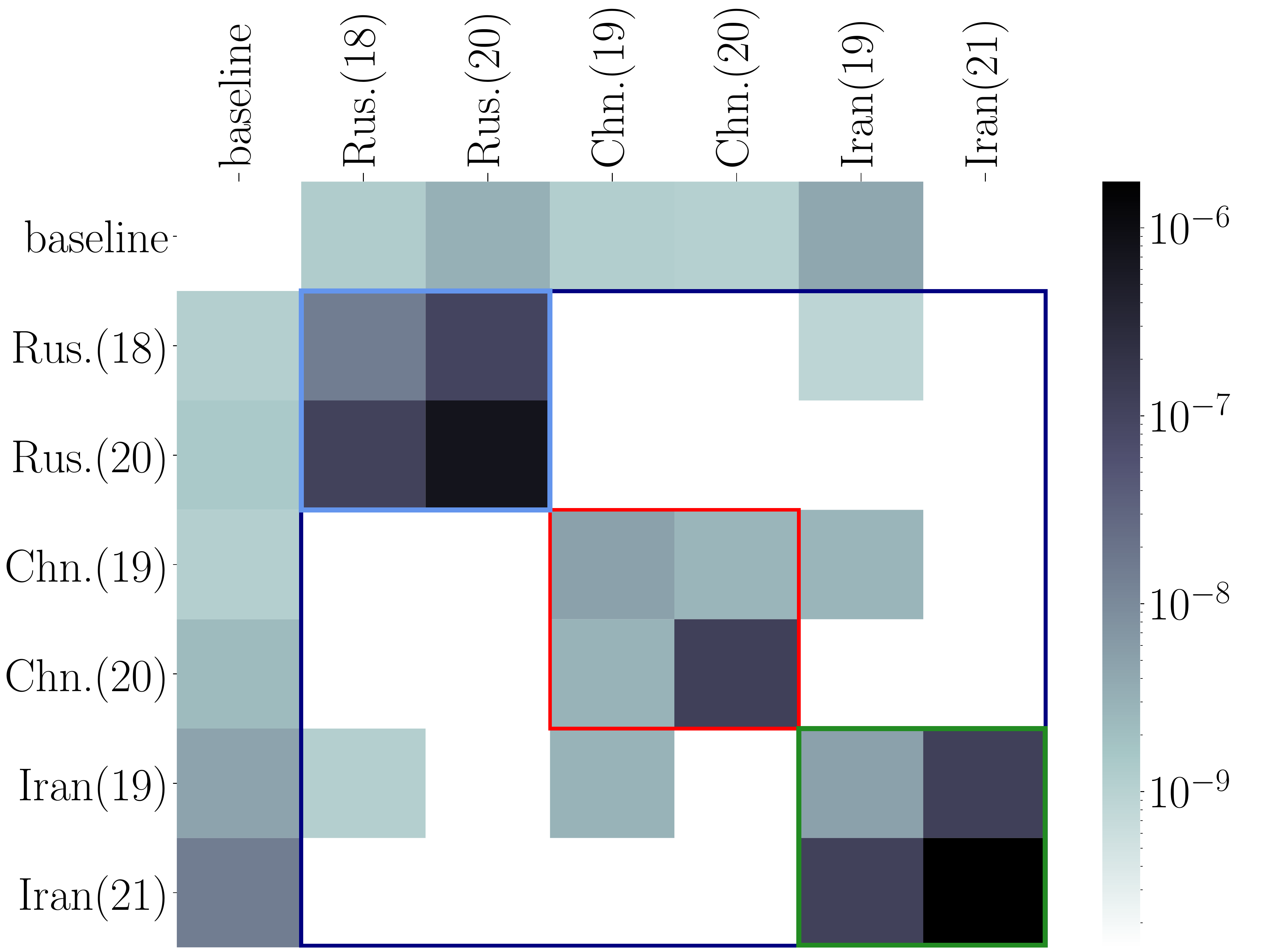}}
        \caption{\centering Frequency near-simultaneous co-URLs ($\tau<1$ min.) between subsets. Counts are normalized by the size of the leading and lagging subsets.} 
    \label{fig:curl0}
    \end{subfigure}

    \vspace{-.3cm}
    \begin{subfigure}[t]{0.45\textwidth}
    \centering
        \makebox[\textwidth][c]{\includegraphics[width=1.2\linewidth]{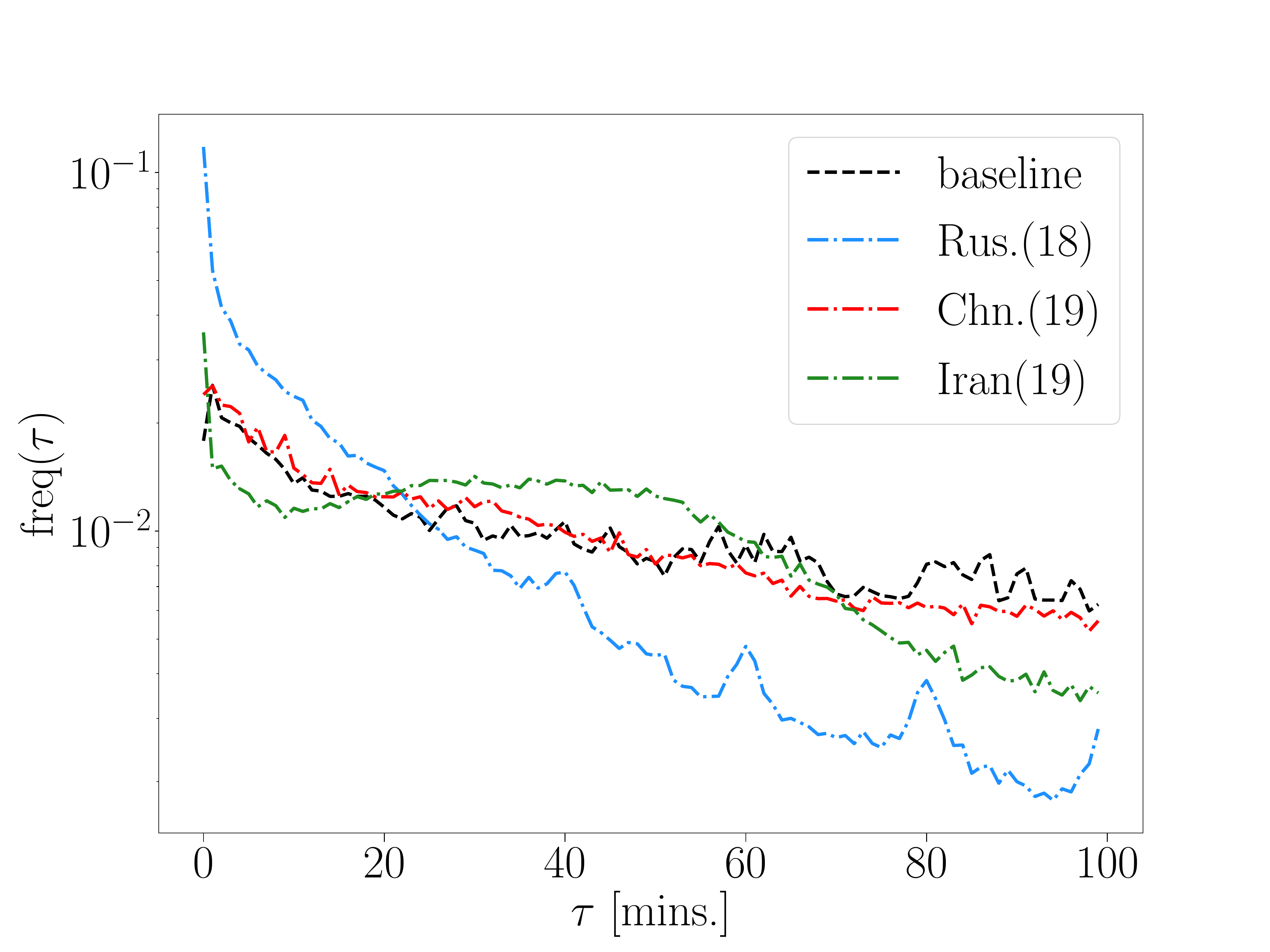}}
        \caption{ \centering Training set co-URL counts as a function of interarrival time $\tau$. Each series corresponds to a diagonal element in Figure \ref{fig:curl} and \ref{fig:curl0} above.} \label{fig:courl_train}
    \end{subfigure}
    \hfill
    \begin{subfigure}[t]{.45\textwidth}
    \centering
        \makebox[\textwidth][c]{\includegraphics[width=1.2\linewidth]{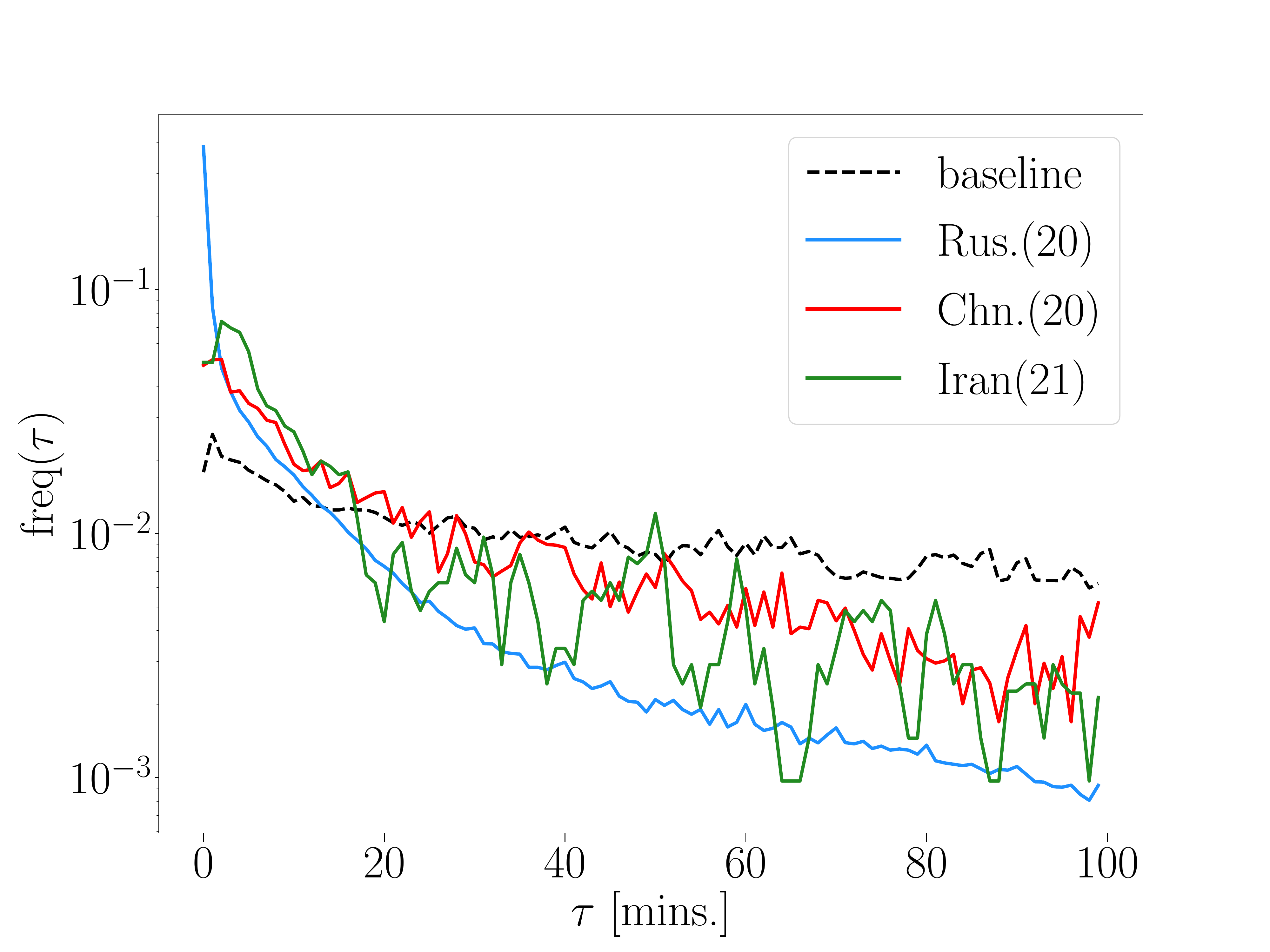}}
        \caption{ \centering Test set co-URL counts as a function of interarrival time $\tau$. Each series corresponds to a diagonal element in Figures \ref{fig:curl} and \ref{fig:curl0} above.} \label{fig:courl_test}
    \end{subfigure}

    \begin{subfigure}[t]{.45\textwidth}
    \centering
        \makebox[\textwidth][c]{\includegraphics[width=1.2\linewidth]{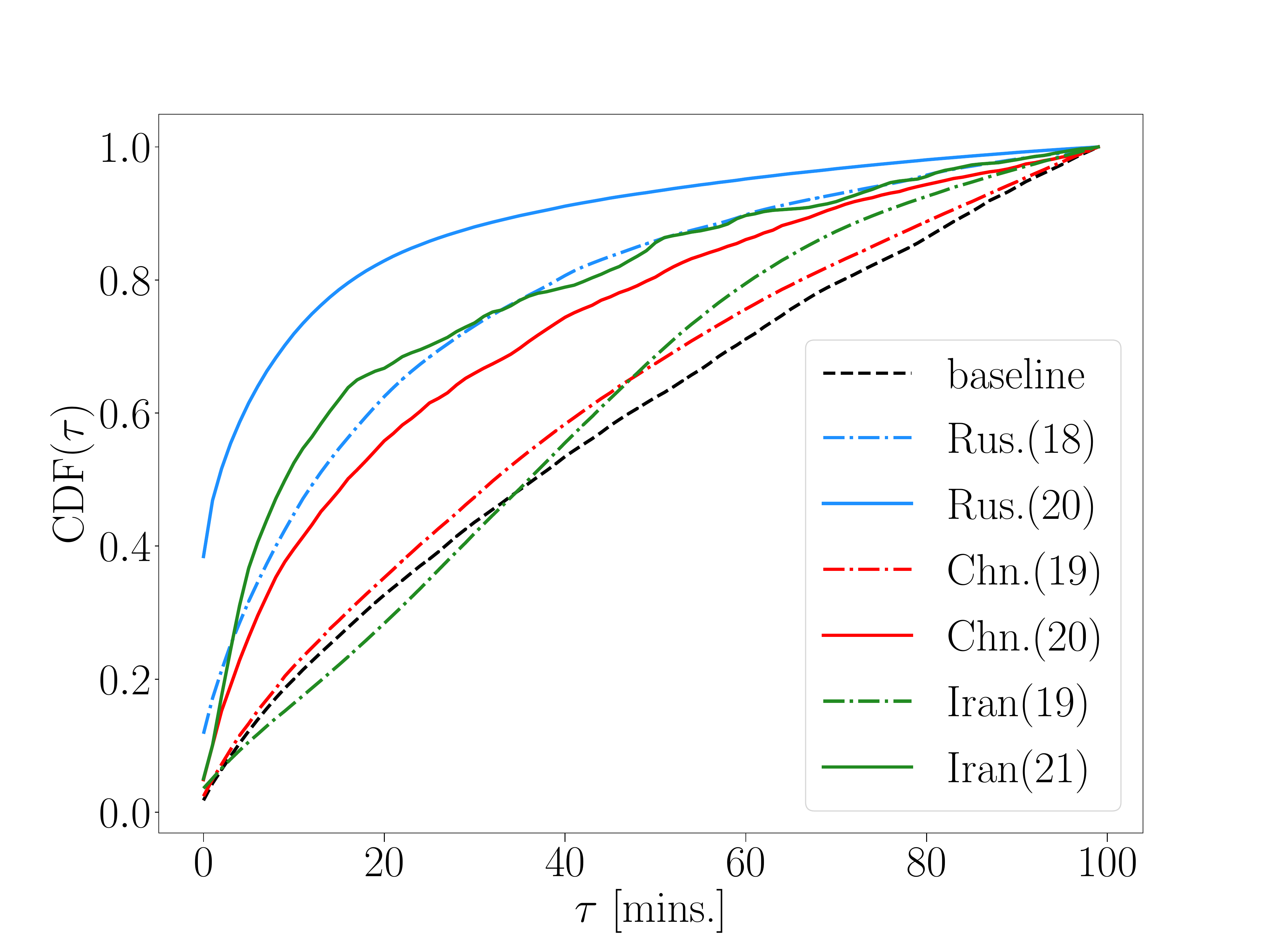}}
        \caption{ \centering Cumulative distribution function of co-URL counts for all campaigns.} \label{fig:cdf}
    \end{subfigure}
    \hfill
    \begin{subfigure}[t]{.45\textwidth}
    \centering
        \makebox[\textwidth][c]{\includegraphics[width=1.3\linewidth]{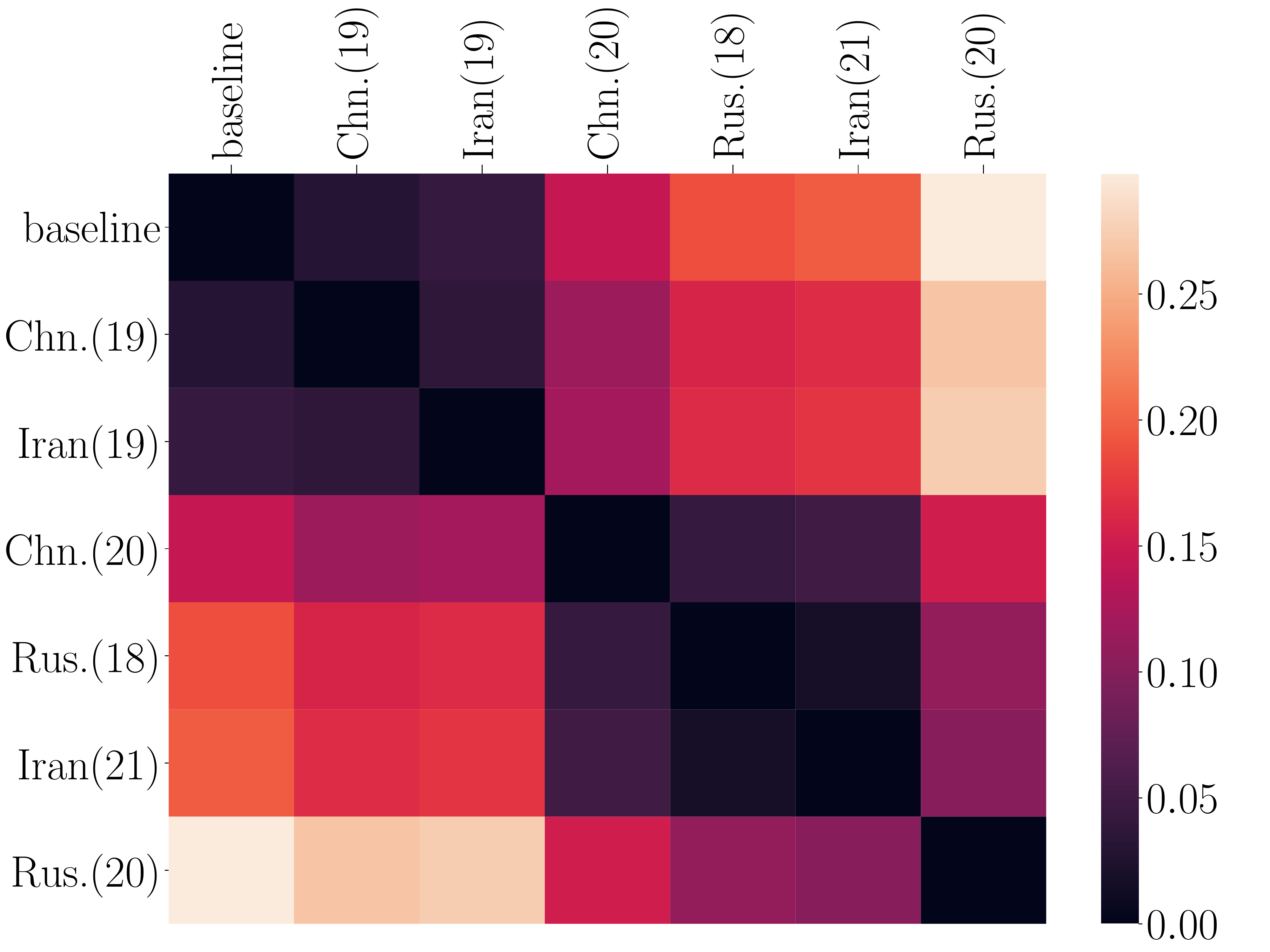}}
        \caption{ \centering Mean absolute distance between CDFs. An increase in near-simultaneous link sharing was observed across operations even as automated detection improved.} 
        \label{fig:MAD}
    \end{subfigure}

    \caption{Summary of co-URL statistics for the 6 IO subsets and baseline.}
    \label{fig:coURL}
\end{figure}

In order to understand intra- and inter-operation patterns of coordination, we report co-URL counts between all campaigns in Figure \ref{fig:coURL}. The 2x2 block pattern of co-URL counts along the diagonal of \ref{fig:curl} and \ref{fig:curl0} suggests that each campaign of a particular origin is actually a continuation of the same underlying operation. Observing how these co-URLs are distributed as a function of interarrival time, we see that the earlier identified training campaigns in \ref{fig:courl_train} have similar distributions of co-URLs as the later identified test campaigns in \ref{fig:courl_train} in some cases. In particular, the campaigns of Chinese and Iranian origin appear to adopt near-simultaneous link sharing later than the Russian campaigns. This can be quantified by calculating the distance between CDFs for each campaign (\ref{fig:cdf}). From \ref{fig:MAD}, we see that the accounts in the Chn.(19) and Iran(19) campaigns appear to use little to no coordination at short timescales compared to the baseline, but the Chn.(20) and Iran(21) campaigns begin to display levels of coordination comparable to the Rus.(18) campaign. The Rus.(20) campaign displays greater levels of coordination than any campaign in our dataset, particularly at short timescales.

\subsection{Feature Importance}
For each model and subtask, we report the best performing graph encoding in each case (G.E. in Table \ref{fig:big_table}). However, it is not clear from these results what the relative importance of each graph-based feature is on model predictions, or the relative importance of content-based to graph-based features. To quantify the relative importance of each feature set, we calculate the mean absolute integrated gradients (IG) for each feature over validation, test, and baseline sets for each subtask (Table \ref{tab:ig_trials} in the Appendix). In Table \ref{tab:ig_agg} we report the aggregated (arithmetic mean) IG values of each feature over all subtasks. We again use MLP as a representative model since it consistently performs well on all subtasks. 

Overall, the net attribution of all graph-based features (\textbf{node2vec}, \textbf{LE}, \textbf{RWPE}, \textbf{NF}) appears to be substantially larger than that of all content-based features (\textbf{domains}), greater by roughly an order of magnitude. Notably, several quantities widely used in network analysis such as Laplacian Eigenmaps (\textbf{LE}), clustering coefficient, and betweenness centrality had a marginal impact on predictions, indicating that they provided little useful information for predictions and, since dropout was employed, that this information was not even redundant with other features. Meanwhile, graph embedding techniques such as \textbf{node2vec} and \textbf{RWPE} enjoy a relatively high utility, having a substantial impact on predictions. This result is not necessarily surprising since \textbf{node2vec} and \textbf{RWPE} essentially act as deep encoders, which can be decoded with high fidelity by deep neural networks (i.e. MLP and GNs, but not by LR and RF). What \textit{is} surprising, on the other hand, is that several simple network quantities—degree, pagerank, and HITS—were as important to predictions as any other single feature. This implies that these quantities encode some information which is complementary to graph embedding techniques, and do so with only a single scalar value.

\begin{table}[!htb]
\hspace*{-1cm}
    \begin{minipage}{.56\linewidth}
      \centering
\begin{center}
\begin{tabular}{l}
      \textbf{(A):  Combined, intra-operation}   \\  \hline 
\end{tabular}
\end{center}
\vspace*{-1mm}

\begin{tabular}{l|l|l|l}
feature             & IG(val.) & IG(test) & IG(base.)  \\  \hline 
\textbf{domains}    & \, $ 9.32 \times 10^{-2}$     &  \, $ 1.06 \times 10^{-1}$   &  \, $ 4.52 \times 10^{-1}$  \\
\textbf{node2vec}   & \, $ 2.11 \times 10^{-1}$     &  \, $ 3.42 \times 10^{-1}$   &  \, $ 1.87 \times 10^{-1}$  \\
\textbf{LE}         & \, $ 9.53 \times 10^{-5}$     &  \, $ 5.32 \times 10^{-4}$   &  \, $ 3.92 \times 10^{-5}$  \\
\textbf{RWPE}       & \, $ 1.18 \times 10^{-1}$     &  \, $ 2.83 \times 10^{-1}$   &  \, $ 3.13 \times 10^{-1}$  \\
\textbf{NF}         & \, $ 7.00 \times 10^{-1}$     &  \, $ 8.68 \times 10^{-1}$   &  \, $ 7.90 \times 10^{-1}$  \\
\,  degree          & \, $ 3.62 \times 10^{-1}$     &  \, $ 3.02 \times 10^{-1}$   &  \, $ 2.60 \times 10^{-1}$  \\
\,  cluster. coef.  & {}\, $ 9.51 \times 10^{-3}$       & {}\, $ 1.56 \times 10^{-3}$   &   {}\, $ 7.34 \times 10^{-3}$  \\
\,  betweenness     & {}\, $ 1.39 \times 10^{-3}$       & \,  $ 4.04 \times 10^{-2}$   &  \, $ 7.83 \times 10^{-2}$  \\
\,  pagerank        & {}\, $ 1.75 \times 10^{-1}$      &  {}\, $ 2.87 \times 10^{-1}$   &   {}\, $ 3.60 \times 10^{-1}$  \\
\,  HITS            & \, $ 1.45 \times 10^{-1}$     &  \, $ 2.20 \times 10^{-1}$   &  \, $ 7.74 \times 10^{-2}$  \\
\end{tabular}
\centering

\end{minipage}%
\begin{minipage}{.56\linewidth}
\centering
\begin{center}
\begin{tabular}{l}
      \textbf{(B):  Combined, inter-operation}   \\  \hline 
\end{tabular}
\end{center}
\vspace*{-1mm}

\begin{tabular}{l|l|l|l}
feature             & IG(val.) & IG(test) & IG(base.)  \\  \hline 
\textbf{domains}    & \, $7.84 \times 10^{-2}$     &  \, $ 9.31 \times 10^{-2}$   &  \, $ 4.50 \times 10^{-1}$   \\
\textbf{node2vec}   & \, $1.94 \times 10^{-1}$     &  \, $ 3.38 \times 10^{-1}$   &  \, $ 1.93 \times 10^{-1}$   \\
\textbf{LE}         & \, $9.98 \times 10^{-5}$     &  \, $ 5.18 \times 10^{-4}$   &  \, $ 4.65 \times 10^{-5}$   \\
\textbf{RWPE}       & \, $1.15 \times 10^{-1}$     &  \, $ 2.87 \times 10^{-1}$   &  \, $ 3.27 \times 10^{-1}$   \\
\textbf{NF}         & \, $6.38 \times 10^{-1}$     &  \, $ 8.36 \times 10^{-1}$   &  \, $ 7.84 \times 10^{-1}$   \\
\,  degree          & \, $3.31 \times 10^{-1}$     &  \, $ 2.97 \times 10^{-1}$   &  \, $ 2.65 \times 10^{-1}$   \\
\,  cluster. coef.  &   {}\, $4.92 \times 10^{-3}$       & \,  $ 3.17 \times 10^{-3}$   &  \, $ 7.22 \times 10^{-3}$ \\
\,  betweenness     &   {}\, $3.65 \times 10^{-3}$      & \, $ 3.97 \times 10^{-2}$   &  \, $ 8.08 \times 10^{-2}$ \\
\,  pagerank        &   {}\, $1.58 \times 10^{-1}$       &  {}\, $ 2.74 \times 10^{-1}$   &   {}\, $ 3.60 \times 10^{-1}$ \\
\,  HITS            & \, $1.35 \times 10^{-1}$     &  \, $ 2.21 \times 10^{-1}$   &  \, $ 7.10 \times 10^{-2}$   \\
\end{tabular}
\end{minipage} 
\caption{Mean absolute integrated gradients (IG) of trained MLPs over features for validation, test, and baseline subsets. For all features, we report the sum of absolute values to avoid cancellation due to conflicting signs. Additionally, we report the IG of each of the five quantities comprising (\textbf{NF}). For IG values of individual subtasks and domains, see Tables \ref{tab:ig_trials} and \ref{table:ig_domains} in the Appendix.}
\label{tab:ig_agg}
\end{table}

\section{Discussion}

\begin{figure}[H]
    \centering
    \includegraphics[width=1.\textwidth]{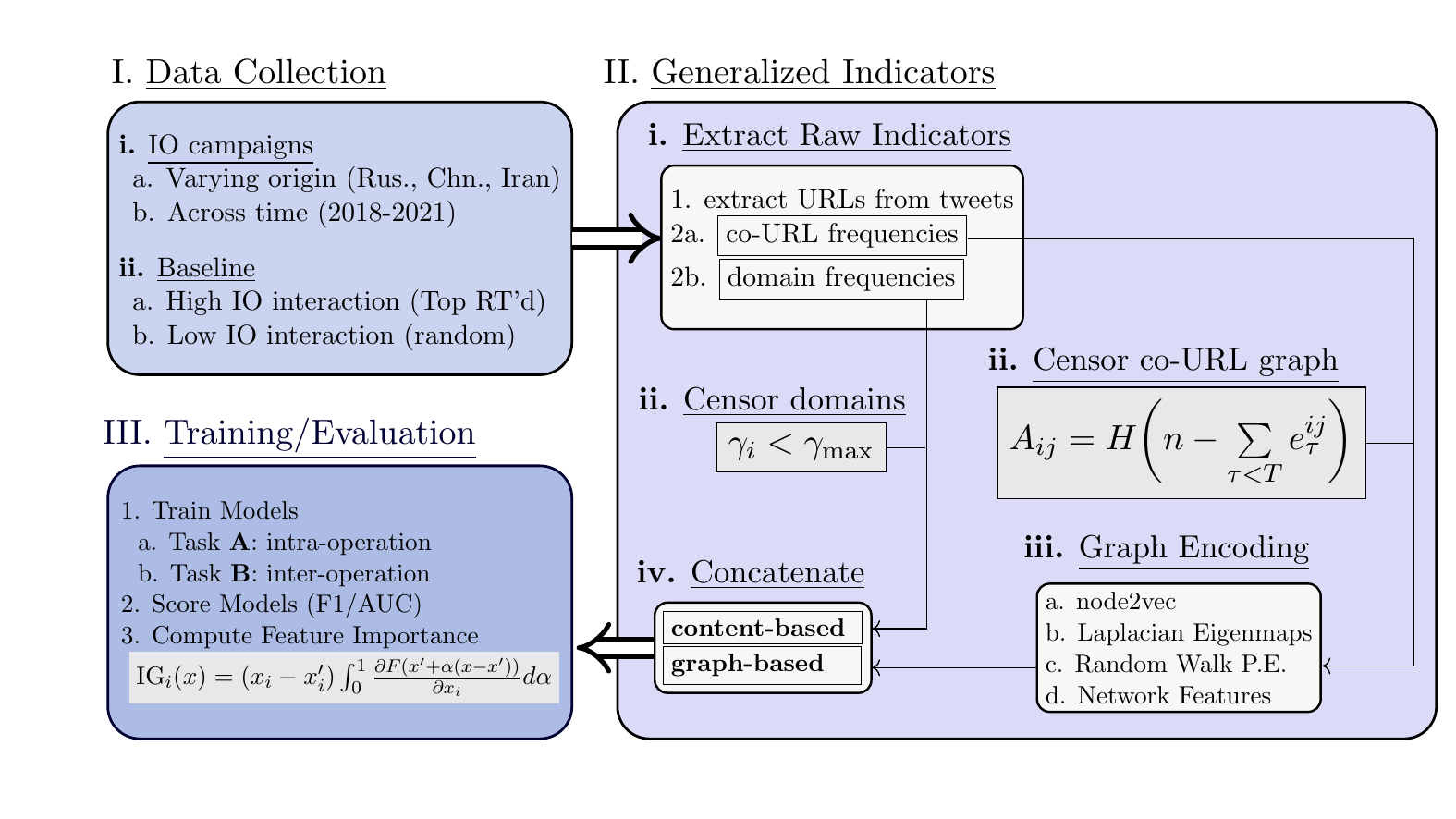}
    \caption{Illustration of the proposed inductive learning framework for the detection of information operations: \textbf{I.}) Collect IO data spanning various operations and time periods, as well as a baseline that interacts with the IO to varying degrees; \textbf{II.}) Extract and censor raw content-based and graph-based indicators and encode signatures of coordination via graph learning; \textbf{III.}) Evaluate model performance on tasks requiring generalization and determine the most important indicators using feature attribution.  $H(\cdot)$ is the Heaviside step function. }
    \label{fig:flow}
\end{figure}

Constraining influence operations is an ongoing challenge that will require continued advancement of detection capabilities in order to counter novel operations—particularly as they adopt powerful AI technologies. In particular, detection methods which go beyond established transductive methodologies and can identify novel campaigns in an inductive manner will be critical. Here we have examined the systematic application of generalized indicators and graph learning techniques, demonstrating a framework in Figure \ref{fig:flow} which enhances detection coverage. Furthermore, this framework is broadly applicable to detecting manipulation on social media, and naturally complements detection using technical indicators identified in transductive methodologies. 

Overall, the most effective approaches utilized: (1) a fairly large content-based feature set (approximately 2000-2500 domains) with fairly stringent removal threshold applied ($\gamma_{\textrm{max}} \approx 0.5$); (2) a broad range of graph-encoding features, particularly \textbf{node2vec}, \textbf{RWPE}, degree, pagerank, and HITS; (3) a deep neural architecture. In other words, MLP and the three GNs outperformed LR and RF on every out-of-sample subtask. On the in-sample prediction tasks, as quantified by F1(val) of tasks \textbf{A1}, \textbf{A2}, and \textbf{A3}, RF actually outperforms all of the deep models.  It appears that in this case RF was simply able to memorize patterns specific to the training set, as it fails to generalize to the test set. For task \textbf{A}, this failure is corresponds $\sim$5 point decrement on F1(test) and AUC(test) relative to the deep models.  On task \textbf{B}, for which predictions had to be made by generalizing across campaigns, an even larger decrement of 10-15 points is observed across all metrics. 

Among deep models, MLP consistently achieved high performance across all subtasks, achieving the highest AUC(test) in all cases but one. However, on all out-of-sample F1 scores one or more GNs outperformed MLP in all cases. This indicates that while GNs perform well at the decision boundary for classification (the n{\"a}ive boundary $\alpha$=0.5 in all cases), misclassifications were by a greater margin than for MLP. This is possibly an indication that while the increased expressive power of GNs was beneficial for classifications on average, this could lead to even further errors on accounts which were the most difficult to classify.

While we have outlined several specific approaches for selecting feature sets and models which can generalize across campaigns, there are a number of improvements which would likely further this work. First, there are presumably other sets of generalized indicators which may be in aid in out-of-sample identification of IO accounts. The first is text data, which we did not investigate. On one hand, text embeddings which encode specific narratives or ideologies could presumably provide useful information which is not necessarily specific to a particular IO campaign. But on the other hand, the length of text shared and the prevalence of text varies greatly among accounts even with a single platform, and even more so from platform to platform. For content-based features overall, a unified approach for encoding the content and semantics of text, images, audio and video would be ideal, since focusing on a specific form of content could lead to blind spots. For example, the multi-modal encoders used by the multi-task agent Gato\cite{gato2022a} or the GPT-4 system\cite{openai2023gpt4} could allow for training on text posts and making inferences on video posts, and so on. 

Though the co-URL is a effective and robust tool for quantifying coordination, there are many other edge-wise features $e_{ij}$ which quantify pair-wise relationships in a graph. In particular, several graph-based measures which quantify similarity could add useful information: node-similarity measures based on nearest neighbors such as common neighbors, Jaccard Index, Adamic Adar, and preferential attachment coefficients; as well as path based measures such as shortest path lengths, Katz measure, and hitting time. Other similarity measures can be derived from graph learning measures by applying various distance metrics such as $L_p$, cosine, and Sørensen–Dice distances to pairs of graph encodings. Fortunately, message passing graph networks provide a natural way to incorporate similarity measures (or any edge-wise features) into predictions, making this type of extension straight forward.

Although we attempted to present a range of graph networks—convolutional, shallow message-passing, and deep message-passing graph networks—there are myriad design dimensions of graph networks which we did not explore. Among these are more advanced sampling strategies, attention mechanisms, and various message passing architectures. However, the results in this study are adequate to suggest that both graph learning and graph networks will be an indispensable tool for detecting IO into the future.

Finally, we examined a ``hard" measure of coordination, the co-URL, in this paper. There could in the future, however, be softer forms of coordination which evade detection. For example, different URLs could lead to semantically or literally identical content, which would not be measured as coordination by our current approach. To hedge against this possibility, one could encode the content of URLs as embeddings and define coordination as a function of the distance between embeddings. This would generalize the current co-URL approach in which we implicitly assign identical URLs distance 0 and distinct URLs distance $\infty$. This is yet another indication of the utility of multi-modal content encoders in future influence detection efforts.

In summary, we have demonstrated an inductive approach to detecting IO which allow for continued utility into the future and generalization capacity across campaigns, enabling identification beyond technical indicators identified by transductive methodologies. We have illustrated how specific content- and graph-based features realize these objectives, as well as how one can systematically identify these features. Finally, we have identified several refinements of the current approach, enabling continued advancement in the automated detection of IO even as these campaigns continue to evolve.

\section{Methods}

\subsection{Data Collection and Inclusion Criteria}

For the purpose of evaluating intra- and inter-operation generalization of machine learning models, we selected IO of several origins (Russia, China, and Iran) for which there were significant campaign sub-networks identified at different times. To this end, six Russian, Chinese, and Iranian origin campaigns identified by Twitter between 2018 and 2021 were suitable. Another important aspect considered was the availability of baseline accounts which both interacted with the IO (to provide adequate coordination measures) and remained independent of IO (to reduce bias in frequency measures). To this end the baselines of \cite{Alizadeh2020,Vargas2021} were used ($88.5\%$ of baseline accounts), in addition to 1129 accounts ($11.5\%$ of baseline accounts) highly interacted with by IO, as measured by co-URLs. Additionally, the 1129 high-interaction accounts were sampled at various maximum follower thresholds ($n = 10^2, 10^3,$ and $10^4$) since accounts with many followers ($n>10^4$) were disproportionately interacted with by IO accounts. This resulted in an aggregate baseline which was highly connected to the IO and yet provided broad coverage of various types and sizes of accounts.

For this study, we focus specifically on IO accounts which displayed reasonably organic patterns of sharing, which evaded detection for some amount of time, and which could have reasonably had some impact on public discourse. We therefore selected from the initial data set accounts which: (1) were active for at least 3 months; (2) had at least 300 Tweets; (3) had at least 200 URL shares; (4) shared at least 5 unique domains; (5) had at least 10 co-URLs with at least 2 neighbors.

\subsection{Extracting raw indicators}
To obtain features from the raw tweet data, we expand all shortened URLs contained in tweets using the URLExpander library \cite{URLexp2018}. Then, to obtain domain counts for each account, we use the tldextract library to extract the domain of each URL tweeted by an account. We then generate node-wise and edge-wise features from the raw URLs and extracted domains:

\begin{itemize}
    \item[] \underline{Edge features}: As a measure of coordination between accounts, we compute the interarrival time between shares of the same URL (co-URLs) and bin the results into 1 minute intervals to obtain a vector of co-URL frequencies between all pairs of accounts. Denoting the interarrival time as $\tau$, the co-URL count between the $i$th and $j$th account in the interarrival window $\tau-1 < t \leq \tau$ is then denoted $e^{ij}_{\tau}$. 
    \item[] \underline{Node features}: We use the raw counts of the most frequently shared top-level domains (e.g.,, cnn.com, youtube.com, nytimes.com) from each account in the composite dataset. To avoid having the models simply memorize domains which are specific to a particular IO, we censor domains where more than $\gamma_{\textrm{max}}$ of occurrences of the domain originate from the IO training set relative to the baseline set. For example, riafan.ru, histantv.com, and tel-avivtimes.com are censored by this method for any $\gamma_{\textrm{max}}<10^4$ since each of these domains originate at least $10^4$ more frequently from IO accounts than from the baseline. See Appendix E for extended examples of censored domains.
    \item[] \underline{Graph Encoding}:  From a censored co-URL graph we compute three graph embeddings—node2vec (dim$=128$), Laplacian Eigenmaps (dim$=50$), Random Walk Positional Encodings (dim$=50$), and several network statistics (degree, clustering coefficient, betweenness centrality, PageRank and HITS). We define the concatenation of all graph-based features the graph encoding.
\end{itemize}


\subsection{Content-based generalized indicators}
Following transductive methodologies, previous work has enabled the rapid detection of IO which attempt to propagate specific domains containing fake news, propaganda, and malware\cite{GraphikaDeepFake,FB-Q1-2023,FB-Q2-2022}. Accounts sharing these domains, particularly in a coordinated manner, are now routinely identified and removed by mainstream platforms. To identify influence efforts beyond these more flagrant indicators, we investigate domains which are commonly shared and yet may be useful indicators of IO activity. In order to quantify the extent to which a particular domain is either common to some baseline users or specific to an IO, we define the relative frequency for the $i$th domain in the IO training set ($y=1$) relative to a baseline set ($y=0$) as

\begin{equation}
    \gamma_i = \frac{\textrm{tf}(i,y=1)}{\textrm{tf}(i,y=0)} 
\end{equation}
where the domain term frequencies for the $i$th domain are
\begin{equation}
\begin{aligned}
    \textrm{tf}(i,y=1) = \frac{f_{i}^{(y=1)}}{\sum_{i}f_{i}^{(y=1)}} \\
    \textrm{tf}(i,y=0) = \frac{f_{i}^{(y=0)}}{\sum_{i}f_{i}^{(y=0)}} 
\end{aligned}
\end{equation}
and $f_i^{(y=1,0)}$ are the raw counts of the $i$th domain in the IO and baseline training sets. We then {censor} any domains which exceed a threshold $\gamma_{\textrm{max}}$ such that we remove domains which are specific to the IO training set with variable stringency. Particular choices of $\gamma_{\textrm{max}}$ can censor domains which only appear in the baseline set ($\gamma_{\textrm{max}}$ = 0), appear in the IO set no more than parity ($\gamma_{\textrm{max}}$ = 1), or which appear only in the IO set ($\gamma_{\textrm{max}}$ = $\infty$). Additionally, we retain only a select number of the censored domains, $k_{\textrm{top}}$, the top-$k$ domains when sorted in descending order by absolute frequency. We can then vary the stringency and minimum prevalence of our content-based feature set with $\gamma_{\textrm{max}}$ and $k_{\textrm{top}}$, respectively, in order to investigate the effect of content censorship on generalization. Moreover, at less stringent thresholds ($\gamma_{\textrm{max}}>1$), we can observe the effect of directly including technical indicators of previous campaigns used in transductive methodologies.

\subsection{Graph-based generalized indicators}

Coordination by IO on social platforms has taken many forms, including mass spamming, mass reporting, and coordinated content sharing. Among these tactics, coordinated content sharing has perhaps been the most widely observed, and is the chief tactic employed by many campaigns. In particular, many takedown efforts have used near-simultaneous co-URL sharing as the primary means of both identifying and substantiating coordinated inauthentic behaviour. In addition to the ubiquity of co-URLs across a variety of campaigns and platforms, they also have the appealing properties that they are agnostic to the specific content shared, are easily defined across platforms, and automatically imply a graph structure between accounts. Furthermore, each co-URL has an associated time between shares, the \textit{interarrival time}, whose distribution can provide further insight into coordinated activity between accounts (see Figure \ref{fig:coURL} for examples). 

While near-simultaneous co-URLs are a useful indicator for automated detection of IO activity, this behaviour is not guaranteed to persist, particularly for campaigns with high operational security (i.e., those which closely mimic authentic users). Therefore, generalized indicators of coordination should incorporate a broader time frame within which future campaigns are likely to operate. In particular, we utilize co-URLs with interarrival times from $\tau=0$ to 100 minutes. This time frame includes near-simultaneous sharing (<1 min.), the majority of retweets (<20 min.\cite{RT20mins}), and the median half-life of tweet views ($\sim$80 mins. \cite{2023halflife}). We denote the number of co-URLs with interarrival times $ \tau-1 < t \leq \tau$ as $e^{ij}_{\tau}$, and the composite co-URL vector as $e_{ij} = \{ e^{ij}_1, \ldots, e^{ij}_{100}\}$.

The graph structure implied by co-URLs, however, cannot be used directly by machine learning models to make node-wise inferences. Two approaches for utilizing graph structured data in machine learning applications are to: (i) learn unsupervised feature vectors for each node in the graph (\textit{graph embedding}); and/or (ii) define graph operators which systematically aggregate data over the graph at each layer in a neural network. Both of these techniques can be referred to collectively as \textit{graph learning}\cite{GRL_chami}, and neural networks utilizing graph operators as \textit{graph networks}.

Graph networks can utilize co-URL data in ways which may or may not directly make use of near-simultaneous link sharing behaviour, thereby offering varying degrees of generalization capacity. For example, one can define a graph which censors near-simultaneous link sharing as follows: assign $A_{ij} = 1$ if and only if two accounts share at least $n$ URLs with interarrival times less than $T$, or in mathematical notation
\begin{equation} \label{eq:graph}
    A_{ij} = H\bigg(n-\sum\limits_{\tau<T} e^{ij}_{\tau} \bigg)
\end{equation}
 where $H(\cdot)$ is the Heaviside step function. This definition equally counts the contribution of all co-URLs with interarrival times less than $T$, thereby censoring any near-simultaneous behavior while still allowing a rigorous threshold for coordination. One can then define graph operators, such as GCN\cite{kipf2016}, in terms of this censored graph. A standard way of directly using vector-valued edge features such as co-URLs, on the other hand, is within a message passing\cite{MPNN2017,RWPE} framework. Message passing defines graph operators directly as a function $\phi(e_{ij})$ of the edge-wise feature vectors (i.e., allowing predictions to be made directly using near-simultaneous co-URLs $e^{ij}_{1}$). In order to understand the generalization capacity of graph networks with varying degrees of graph censorship, we implement three graph networks as follows: GCN, which utilizes only the censored graph;  MP-GCN(s), a message passing variant of the base GCN architecture with a shallow message passing function $\phi = L$, where $L$ is a linear operator; and MP-GCN, which uses the more common deep message passing function $\phi = f$, where $f$ is a neural network. Comparing the performance of these three architectures allows us to examine the effect of graph censorship, as well as compare different graph network architectures in identifying IO.

\subsection{Graph Encoding}
In order to understand the utility of different types of graph-based features (from network analysis to graph learning) as well as the utility of specific features, we incorporate several candidate quantities in a node-wise feature vector which we call the graph encoding. Due to the asymmetric nature of our dataset (co-shares of content \textit{by} IO accounts are visible in the dataset, but co-shares \textit{of} IO account content have been removed by Twitter) we treat all graph quantities in an undirected manner by setting $e_{ij} \leftarrow e_{ij} + e_{ji}$. All graph-based features are derived from an undirected graph computed from the co-URL vectors as in 
Equation \ref{eq:graph} where we select thresholds of $n=10$ and $T=15$ to censor the graph. While more stringent $n$ would produce a more robust graph, we find that further reducing the number of edges rapidly disjoints the graph, making graph learning techniques infeasible. The temporal threshold $T=15$ minutes represents a window in which IO could coordinate effectively and yet avoid detection, while also censoring near-simultaneous link sharing.

Graph representation learning, including graph embedding algorithms such as node2vec\cite{node2vec}, originated as an effort to automate the feature engineering process for graph prediction tasks such as node classification and link prediction. From a modern perspective, graph embedding techniques are  unsupervised methods which allow one to systematically assign relational, functional, and structural information to each node in a graph. This information can greatly improve the performance  of deep learning models, with or without graph operators, on graph prediction tasks. We choose three graph embedding algorithms for our purpose here: (1) node2vec, which encodes neighborhood information of nodes into dense embeddings; (2) Laplacian Eigenmaps, a non-linear spectral embedding technique which provides a local coordinate system on graphs and effectively encodes clustering within the graph; (3) Random Walk Positional Encoding, which is based on the graph diffusion operator and uniquely assigns node embeddings based on the $k$-hop topological neighborhood of each node. Each of these approaches, in principle, encode different aspects of graph topology and therefore can provide predictive utility independent of one another. In each case, the dimensions of the embeddings are chosen such that further increases yield no benefit to performance across models.

While graph embeddings are a sensible method of encoding topological information for predictive tasks, they do not necessarily preclude the utility of conceptually similar network analysis quantities. To this end we include several quantities which encode relational, functional, and structural information of graphs in our graph encoding: (1) degree, which for undirected graphs is simply the number of directly adjacent neighbors of each node; (2) clustering-coefficient, which quantifies the local clustering of each node as the amount of closure between the neighbors of each node; (3) betweenness centrality, a centrality measure quantifying the extent to which a node facilitates connection within the graph via shortest paths; (4) pagerank, a centrality measure which ranks nodes according to their relative importance within a network; (5) HITS, which also ranks nodes according to relative importance but assigns two scores quantifying the extent to which a node connects the graph (hub score) and is of relative importance within the graph (authority score). For undirected graphs, the hub and authority scores of HITS are identical. 

\subsection{Graph Networks}

Given the node-wise and edge-wise data in our feature set, there are several GN architectures which are possible choices for the predictive task at hand. We sample several architectures of increasing expressive power such that we can compare the utility of different GN design choices and degrees of graph censorship.  In particular, we perform ablation on message-passing rules for encoding the co-URL vectors $e^{ij}_{\tau}$ in order to compare various uses of this feature set. 

In general, a graph network can be written as the series of operations 
\begin{flalign}
    &&h_i^{(l+1)} &= W^{(l)} h_i^{(l)} + b^{(l)}&&&&&&&&&&& \mbox{(affine transformation)}\\
    &&h_i^{(l+1)} &= \underset{j \in \mathcal{N}(i)}{\textrm{AGG}}\bigg(h_j^{(l+1)}\bigg)&&&&&&&&&&& \mbox{(feature aggregation)}\\
    &&h_i^{(l+1)} &= \sigma\bigg( h_i^{(l+1)} \bigg)&&&&&&&&&&& \mbox{(non-linearity)}
\end{flalign}
where the set $\mathcal{N}(i)$ indicates the neighborhood of the $i$th node where $A_{ij}=1$. The simplest graph network that we employ is a spectral GN, the popular Graph Convolutional Network (GCN), with layers defined by the feature aggregation function\cite{kipf2016}
\begin{flalign}
        &&\underset{j \in \mathcal{N}(i)}{\textrm{AGG}} = \sum_{j \in \mathcal{N}(i)} \frac{1}{\sqrt{d_i d_j}} h_j^{(l+1)} &&&&&&&&&&&&& \mbox{(GCN)}
\end{flalign}
where $d_i$ is the degree of the $i$th node.

There are a number of ways in which message passing rules can be defined, but for graph networks one typically defines the message passing rule as
\begin{equation}
    m_{ij}^{(l+1)} = \phi\bigg(h_i^{(l)},h_j^{(l)},e_{ij}\bigg)
\end{equation}
where the message passing function $\phi(\cdot)$ can take as input both nodewise features $h_i^{(l)}$ and edgewise features $e_{ij}$. We then incorporate these messages into the aggregation step as 
\begin{flalign}
    h_i^{(l+1)} &= \underset{j \in \mathcal{N}(i)}{\textrm{AGG}}\bigg(h_j^{(l+1)},m_{ij}^{(l+1)}\bigg) 
\end{flalign}
We employ two message passing rules to encode the co-URL vector, the first of which is a shallow message passing rule which defines our MP-GCN(s):
\begin{equation}
    m^{(l+1)}_{ij} = \sigma \bigg(\sum_{\tau} w^{(l)}_{\tau} e^{ij}_{\tau} \bigg).
\end{equation}
The second rule utilizes a neural message passing function, implemented as an $L$ layer perceptron which defines our MP-GCN:
\begin{equation}
\begin{aligned}
    a_{ij}^{(k+1)} &= \sigma\bigg(W^{(k,l)}a_{ij}^{(k)} + b^{(k,l)}\bigg); \\
    m^{(l+1)}_{ij} &= \sigma\bigg(W^{(L)}a_{ij}^{(L)} + b^{(L)}\bigg);
\end{aligned}
\end{equation}
where $W^{(l)}$ and $b^{(l)}$ are the weights and biases of the $l$th layer and $a_{ij}^{(0)} = e^{ij}_{\tau}$. To compare with the base GCN implementation, we insert each message passing rule into the base GCN aggregation function as 
\begin{flalign}
        &&\underset{j \in \mathcal{N}(i)}{\textrm{AGG}} = \sum_{j \in \mathcal{N}(i)} \frac{m_{ij}}{\sqrt{d_i d_j}} h_j^{(l+1)}. &&&&&&&&&&&&& \mbox{(MP-GCN(s)/MP-GCN)}
\end{flalign}
Thus we have performed ablation on the message passing rule over the three GCN architectures.

\subsection{Model Training}
 In the LR and RF implementations we tune all hyperparameters to achieve the best model performance via gridsearch. In the MLP and the three GCN variants we use the same hyperparameters: two hidden layers of 64 units, and a dropout probability $p=0.5$ applied to all units in the hidden layers. In all message passing layers we apply a dropout probability of $p=0.2$. For all MLP and GCN training we use a binary cross entropy loss and the Adam optimizer with a learning rate of $10^{-4}$. 

 \subsection{Integrated Gradients}
 Integrated gradients\cite{IntegratedGradients} is an axiomatic attribution method for deep neural networks. Mathematically, the IG of a function $F(x)$ with respect to the $i$th component of an input $x$ and a baseline $x'$ is
 \begin{equation}
     \textrm{IntegratedGradient}_i(x) = (x_i - x_i') \int_0^1 \frac{\partial F(x' + \alpha (x-x'))}{\partial x_i} d\alpha
 \end{equation}
where $\alpha$ parameterizes a straight line path from $x'$ to $x$. This method provides a more robust attribution of predictions to specific features than directly evaluating the product of the gradient and feature value
\begin{equation}
    \textrm{Attr}_i(x) = x_i \frac{\partial F}{\partial x_i}
\end{equation}
which has historically been a popular attribution method. When using IG, one selects a baseline where the model prediction is neutral. Calculating the IGs of each feature for an MLP, there is not an obvious baseline which yields neutral predictions, i.e., where $F(x') = 0.5$. For example, simply choosing the mean or minimum value of each feature over various subsets of the data produces predictions close to 0 or 1. We therefore construct an empirical baseline comprising the subset of all nodes such that $0.4 \leq F(x_j) \leq 0.6$, or within $\pm$0.1 of a neutral prediction. Setting $x'=\langle x_j \rangle$ then yields $F(x') = 0.534\pm0.018$ over all six subtasks, which is approximately neutral while ensuring that no particular feature in the baseline takes on an extreme value (which might be the case if we simply chose $j$ to be the single most neutral prediction).
\subsection{Error propagation of aggregated metrics}
In Figure \ref{fig:gamma_varied}, several performance metrics are aggregated over subtasks by computing their harmonic mean. For each subtask and choice of parameters ($\gamma_{\textrm{max}}$ and $k_{\textrm{top}}$), there is an associated uncertainty for each metric due to their dependence on a random samples of train/validate/test splits in the data. In order to compare aggregated results for different parameter values, we propagate the uncertainties associated with each metric as follows. In general, the harmonic mean can be written
\begin{equation}
    \tilde{x} = \frac{n}{\sum_{i=1}^n x_i^{-1}}
\end{equation}
and the propagated uncertainty (neglecting correlations between $x_i$)
\begin{equation}
    \sigma_{\tilde{x}}^2 = \sum_{i=1}^n \bigg( \frac{\partial \tilde{x}}{\partial x_i}\bigg) \sigma_{x_i}^2.
\end{equation}
The partial derivatives of $\tilde{x}$ with respect to each $x_i$ are
\begin{equation}
        \frac{\partial \tilde{x}}{\partial x_i} = \frac{\tilde{x}^2}{n} \frac{1}{x_i^2} \\
\end{equation}
and the propagated uncertainty is then
\begin{equation}
        \sigma_{\tilde{x}}^2 = \bigg(\frac{\tilde{x}^2}{n}\bigg)^2 \sum_{i=1}^n \bigg( \frac{\sigma_{x_i}}{x_i^2} \bigg)^2.
\end{equation}
Using this result we can better understand different choices of $\gamma_{\textrm{max}}$ and $k_{\textrm{top}}$ shown in Figure \ref{fig:gamma_varied}.

\section*{Code and data availability}
Upon acceptance, code and data will be shared on an individual basis.

\section*{Acknowledgements}

This material is based upon work supported by the Air Force Office of Scientific Research under award number FA9550-20-1-0382.

\bibliography{refs}

\begin{thebibliography}{10}
\urlstyle{rm}
\expandafter\ifx\csname url\endcsname\relax
  \def\url#1{\texttt{#1}}\fi
\expandafter\ifx\csname urlprefix\endcsname\relax\def\urlprefix{URL }\fi
\expandafter\ifx\csname doiprefix\endcsname\relax\def\doiprefix{DOI: }\fi
\providecommand{\bibinfo}[2]{#2}
\providecommand{\eprint}[2][]{\url{#2}}

\bibitem{Broniatowski2018}
\bibinfo{author}{Broniatowski, D.~A.} \emph{et~al.}
\newblock \bibinfo{journal}{\bibinfo{title}{Weaponized health communication:
  Twitter bots and russian trolls amplify the vaccine debate}}.
\newblock {\emph{\JournalTitle{American Journal of Public Health}}}
  \textbf{\bibinfo{volume}{108}} (\bibinfo{year}{2018}).

\bibitem{zannettou2019let}
\bibinfo{author}{Zannettou, S.} \emph{et~al.}
\newblock \bibinfo{title}{Who let the trolls out? towards understanding
  state-sponsored trolls}.
\newblock In \emph{\bibinfo{booktitle}{Proceedings of the 10th ACM Web
  Science}}, \bibinfo{pages}{353--362} (\bibinfo{year}{2019}).

\bibitem{zhou2019elites}
\bibinfo{author}{Zhou, Y.}, \bibinfo{author}{Dredze, M.},
  \bibinfo{author}{Broniatowski, D.~A.} \& \bibinfo{author}{Adler, W.~D.}
\newblock \bibinfo{journal}{\bibinfo{title}{Elites and foreign actors among the
  alt-right: The gab social media platform}}.
\newblock {\emph{\JournalTitle{First Monday}}}  (\bibinfo{year}{2019}).

\bibitem{linvill2020troll}
\bibinfo{author}{Linvill, D.~L.} \& \bibinfo{author}{Warren, P.~L.}
\newblock \bibinfo{journal}{\bibinfo{title}{Troll factories: Manufacturing
  specialized disinformation on twitter}}.
\newblock {\emph{\JournalTitle{Political Communication}}}
  \textbf{\bibinfo{volume}{37}}, \bibinfo{pages}{447--467}
  (\bibinfo{year}{2020}).

\bibitem{rossetti2023bots}
\bibinfo{author}{Rossetti, M.} \& \bibinfo{author}{Zaman, T.}
\newblock \bibinfo{journal}{\bibinfo{title}{Bots, disinformation, and the first
  impeachment of {US} president {Donald Trump}}}.
\newblock {\emph{\JournalTitle{PloS one}}} \textbf{\bibinfo{volume}{18}},
  \bibinfo{pages}{e0283971} (\bibinfo{year}{2023}).

\bibitem{FB-CIB-2022-China-Russia}
\bibinfo{author}{{Nimmo et al.}}
\newblock \bibinfo{journal}{\bibinfo{title}{Taking down coordinated inauthentic
  behavior from {Russia and China}}}.
\newblock {\emph{\JournalTitle{{Meta Newsroom}}}}  (\bibinfo{year}{2022}).

\bibitem{FB-Q2-2022}
\bibinfo{author}{{Nimmo et al.}}
\newblock \bibinfo{journal}{\bibinfo{title}{Quarterly adversarial threat report
  ({Q2})}}.
\newblock {\emph{\JournalTitle{{Meta Newsroom}}}}  (\bibinfo{year}{2022}).

\bibitem{FB-Q3-2022}
\bibinfo{author}{{Meta}}.
\newblock \bibinfo{journal}{\bibinfo{title}{Quarterly adversarial threat report
  ({Q3})}}.
\newblock {\emph{\JournalTitle{{Meta Newsroom}}}}  (\bibinfo{year}{2022}).

\bibitem{FB-Q4-2022}
\bibinfo{author}{{Nimmo et al.}}
\newblock \bibinfo{journal}{\bibinfo{title}{Quarterly adversarial threat report
  ({Q4})}}.
\newblock {\emph{\JournalTitle{{Meta Newsroom}}}}  (\bibinfo{year}{2023}).

\bibitem{FB-Q1-2023}
\bibinfo{author}{{Meta}}.
\newblock \bibinfo{journal}{\bibinfo{title}{Quarterly adversarial threat report
  ({Q1})}}.
\newblock {\emph{\JournalTitle{{Meta Newsroom}}}}  (\bibinfo{year}{2022}).

\bibitem{Twitter-June-2020}
\bibinfo{author}{{Twitter Safety}}.
\newblock \bibinfo{journal}{\bibinfo{title}{Disclosing networks of state-linked
  information operations we’ve removed}}.
\newblock {\emph{\JournalTitle{{Twitter Blog}}}}  (\bibinfo{year}{2020}).

\bibitem{Twitter-Feb-2021}
\bibinfo{author}{{Twitter Safety}}.
\newblock \bibinfo{journal}{\bibinfo{title}{Disclosing networks of state-linked
  information operations}}.
\newblock {\emph{\JournalTitle{{Twitter Blog}}}}  (\bibinfo{year}{2021}).

\bibitem{etudo2019facebook}
\bibinfo{author}{Etudo, U.}, \bibinfo{author}{Yoon, V.~Y.} \&
  \bibinfo{author}{Yaraghi, N.}
\newblock \bibinfo{title}{From facebook to the streets: Russian troll ads and
  black lives matter protests}.
\newblock In \emph{\bibinfo{booktitle}{Proceedings of the 52nd Hawaii
  International Conference on System Sciences}} (\bibinfo{year}{2019}).

\bibitem{reddit2019bots}
\bibinfo{author}{Hurtado, S.}, \bibinfo{author}{Ray, P.} \&
  \bibinfo{author}{Marculescu, R.}
\newblock \bibinfo{title}{Bot detection in reddit political discussion}.
\newblock In \emph{\bibinfo{booktitle}{Proceedings of the Fourth International
  Workshop on Social Sensing}}, \bibinfo{pages}{30–35}
  (\bibinfo{year}{2019}).

\bibitem{zannettou2020characterizing}
\bibinfo{author}{Zannettou, S.} \emph{et~al.}
\newblock \bibinfo{title}{Characterizing the use of images in state-sponsored
  information warfare operations by russian trolls on twitter}.
\newblock In \emph{\bibinfo{booktitle}{Proceedings of the International AAAI
  Conference on Web and Social Media}} (\bibinfo{year}{2020}).

\bibitem{MIT-LL}
\bibinfo{author}{Smith, S.~T.} \emph{et~al.}
\newblock \bibinfo{journal}{\bibinfo{title}{Automatic detection of influential
  actors in disinformation networks}}.
\newblock {\emph{\JournalTitle{Proceedings of the National Academy of
  Sciences}}} \textbf{\bibinfo{volume}{118}} (\bibinfo{year}{2021}).

\bibitem{Alizadeh2020}
\bibinfo{author}{Alizadeh, M.}, \bibinfo{author}{Shapiro, J.~N.},
  \bibinfo{author}{Buntain, C.} \& \bibinfo{author}{Tucker, J.~A.}
\newblock \bibinfo{journal}{\bibinfo{title}{Content-based features predict
  social media influence operations}}.
\newblock {\emph{\JournalTitle{Science Advances}}} \textbf{\bibinfo{volume}{6}}
  (\bibinfo{year}{2020}).

\bibitem{TwitterGDL}
\bibinfo{author}{Monti, F.}, \bibinfo{author}{Frasca, F.},
  \bibinfo{author}{Eynard, D.}, \bibinfo{author}{Mannion, D.} \&
  \bibinfo{author}{Bronstein, M.~M.}
\newblock \bibinfo{journal}{\bibinfo{title}{Fake news detection on social media
  using geometric deep learning}}.
\newblock {\emph{\JournalTitle{CoRR}}}
  \textbf{\bibinfo{volume}{abs/1902.06673}} (\bibinfo{year}{2019}).

\bibitem{Vargas2021}
\bibinfo{author}{Vargas, L.}, \bibinfo{author}{Emami, P.} \&
  \bibinfo{author}{Traynor, P.}
\newblock \bibinfo{journal}{\bibinfo{title}{On the detection of disinformation
  campaign activity with network analysis}}.
\newblock {\emph{\JournalTitle{Proceedings of ACM SIGSAC Conference on Cloud
  Computing Security}}}  (\bibinfo{year}{2020}).

\bibitem{KillChainOO}
\bibinfo{author}{Nimmo, B.} \& \bibinfo{author}{Hutchins, E.}
\newblock \bibinfo{journal}{\bibinfo{title}{Phase-based tactical analysis of
  online operations}}.
\newblock {\emph{\JournalTitle{{Carnegie Endowment for International Peace}}}}
  (\bibinfo{year}{2023}).

\bibitem{KillChainU}
\bibinfo{author}{Pols, P.}
\newblock \bibinfo{journal}{\bibinfo{title}{The unified kill chain}}.
\newblock {\emph{\JournalTitle{{Fox-IT}}}}  (\bibinfo{year}{2017}).

\bibitem{AI_disinfo}
\bibinfo{author}{Sedova, K.} \emph{et~al.}
\newblock \bibinfo{journal}{\bibinfo{title}{{AI} and the future of
  disinformation campaigns, part 1: The richdata framework}}.
\newblock {\emph{\JournalTitle{{Georgetown Center for Security and Emerging
  Technology}}}}  (\bibinfo{year}{2021}).

\bibitem{GraphikaBadRep}
\bibinfo{author}{{Graphika}} \& \bibinfo{author}{{The Stanford Internet
  Observatory}}.
\newblock \bibinfo{journal}{\bibinfo{title}{Bad reputation}}.
\newblock {\emph{\JournalTitle{{Graphika Reports}}}}  (\bibinfo{year}{2022}).

\bibitem{Giglietto2020}
\bibinfo{author}{Giglietto, F.}, \bibinfo{author}{Righetti, N.},
  \bibinfo{author}{Rossi, L.} \& \bibinfo{author}{Marino, G.}
\newblock \bibinfo{journal}{\bibinfo{title}{It takes a village to manipulate
  the media: coordinated link sharing behavior during 2018 and 2019 italian
  elections}}.
\newblock {\emph{\JournalTitle{Information, Communication \& Society}}}
  \textbf{\bibinfo{volume}{23}}, \bibinfo{pages}{867--891}
  (\bibinfo{year}{2020}).

\bibitem{FB2021}
\bibinfo{author}{{Facebook}}.
\newblock \bibinfo{journal}{\bibinfo{title}{Threat report: The state of
  influence operations 2017-2020}}.
\newblock {\emph{\JournalTitle{{Meta Newsroom}}}}  (\bibinfo{year}{2021}).

\bibitem{uprank2010}
\bibinfo{author}{Das~Sarma, A.} \emph{et~al.}
\newblock \bibinfo{journal}{\bibinfo{title}{Ranking mechanisms in twitter-like
  forums}}.
\newblock {\emph{\JournalTitle{Proceedings of the Third ACM WSDM}}}
  (\bibinfo{year}{2010}).

\bibitem{GraphikaDeepFake}
\bibinfo{author}{{Graphika}}.
\newblock \bibinfo{journal}{\bibinfo{title}{Deepfake it till you make it}}.
\newblock {\emph{\JournalTitle{{Graphika Reports}}}}  (\bibinfo{year}{2023}).

\bibitem{SecondaryInfektion}
\bibinfo{author}{{Nimmo et al.}}
\newblock \bibinfo{journal}{\bibinfo{title}{Secondary infektion}}.
\newblock {\emph{\JournalTitle{{Graphika Reports}}}}  (\bibinfo{year}{2020}).

\bibitem{stylegan2}
\bibinfo{author}{Karras, T.} \emph{et~al.}
\newblock \bibinfo{journal}{\bibinfo{title}{Analyzing and improving the image
  quality of stylegan}}.
\newblock {\emph{\JournalTitle{CoRR}}}
  \textbf{\bibinfo{volume}{abs/1912.04958}} (\bibinfo{year}{2019}).

\bibitem{perov2021deepfacelab}
\bibinfo{author}{Perov, I.} \emph{et~al.}
\newblock \bibinfo{journal}{\bibinfo{title}{Deepfacelab: {A} simple, flexible
  and extensible face swapping framework}}.
\newblock {\emph{\JournalTitle{CoRR}}}
  \textbf{\bibinfo{volume}{abs/2005.05535}} (\bibinfo{year}{2020}).

\bibitem{openai2023gpt4}
\bibinfo{author}{OpenAI}.
\newblock \bibinfo{journal}{\bibinfo{title}{Gpt-4 technical report}}.
\newblock {\emph{\JournalTitle{CoRR}}}
  \textbf{\bibinfo{volume}{abs/2303.08774}} (\bibinfo{year}{2023}).

\bibitem{dalle}
\bibinfo{author}{Ramesh, A.} \emph{et~al.}
\newblock \bibinfo{journal}{\bibinfo{title}{Zero-shot text-to-image
  generation}}.
\newblock {\emph{\JournalTitle{CoRR}}}
  \textbf{\bibinfo{volume}{abs/2102.12092}} (\bibinfo{year}{2021}).

\bibitem{IntegratedGradients}
\bibinfo{author}{Sundararajan, M.}, \bibinfo{author}{Taly, A.} \&
  \bibinfo{author}{Yan, Q.}
\newblock \bibinfo{journal}{\bibinfo{title}{Axiomatic attribution for deep
  networks}}.
\newblock {\emph{\JournalTitle{CoRR}}}
  \textbf{\bibinfo{volume}{abs/1703.01365}} (\bibinfo{year}{2017}).

\bibitem{gato2022a}
\bibinfo{author}{Reed, S.} \emph{et~al.}
\newblock \bibinfo{journal}{\bibinfo{title}{A generalist agent}}.
\newblock {\emph{\JournalTitle{Transactions on Machine Learning Research}}}
  (\bibinfo{year}{2022}).

\bibitem{URLexp2018}
\bibinfo{author}{Yin, L.}
\newblock \bibinfo{title}{Smappnyu/urlexpander: Initial release}
  (\bibinfo{year}{2018}).

\bibitem{RT20mins}
\bibinfo{author}{Yin, H.}, \bibinfo{author}{Yang, S.}, \bibinfo{author}{Song,
  X.}, \bibinfo{author}{Liu, W.} \& \bibinfo{author}{Li, J.}
\newblock \bibinfo{journal}{\bibinfo{title}{Deep fusion of multimodal features
  for social media retweet time prediction}}.
\newblock {\emph{\JournalTitle{World Wide Web}}} \textbf{\bibinfo{volume}{24}},
  \bibinfo{pages}{1027--1044} (\bibinfo{year}{2021}).

\bibitem{2023halflife}
\bibinfo{author}{Pfeffer, J.}, \bibinfo{author}{Matter, D.} \&
  \bibinfo{author}{Sargsyan, A.}
\newblock \bibinfo{journal}{\bibinfo{title}{The half-life of a tweet}}.
\newblock {\emph{\JournalTitle{CoRR}}}
  \textbf{\bibinfo{volume}{abs/2302.09654}} (\bibinfo{year}{2023}).

\bibitem{GRL_chami}
\bibinfo{author}{Chami, I.}, \bibinfo{author}{Abu-El-Haija, S.},
  \bibinfo{author}{Perozzi, B.}, \bibinfo{author}{R\'{e}, C.} \&
  \bibinfo{author}{Murphy, K.}
\newblock \bibinfo{journal}{\bibinfo{title}{Machine learning on graphs: A model
  and comprehensive taxonomy}}.
\newblock {\emph{\JournalTitle{Journal of Machine Learning Research}}}
  \textbf{\bibinfo{volume}{23}}, \bibinfo{pages}{1--64} (\bibinfo{year}{2022}).

\bibitem{kipf2016}
\bibinfo{author}{Kipf, T.~N.} \& \bibinfo{author}{Welling, M.}
\newblock \bibinfo{journal}{\bibinfo{title}{Semi-supervised classification with
  graph convolutional networks}}.
\newblock {\emph{\JournalTitle{CoRR}}}
  \textbf{\bibinfo{volume}{abs/1609.02907}} (\bibinfo{year}{2016}).

\bibitem{MPNN2017}
\bibinfo{author}{Gilmer, J.}, \bibinfo{author}{Schoenholz, S.~S.},
  \bibinfo{author}{Riley, P.~F.}, \bibinfo{author}{Vinyals, O.} \&
  \bibinfo{author}{Dahl, G.~E.}
\newblock \bibinfo{journal}{\bibinfo{title}{Neural message passing for quantum
  chemistry}}.
\newblock {\emph{\JournalTitle{Proceedings of Machine Learning Research}}}
  \textbf{\bibinfo{volume}{70}}, \bibinfo{pages}{1263--1272}
  (\bibinfo{year}{2017}).

\bibitem{RWPE}
\bibinfo{author}{Dwivedi, V.~P.}, \bibinfo{author}{Luu, A.~T.},
  \bibinfo{author}{Laurent, T.}, \bibinfo{author}{Bengio, Y.} \&
  \bibinfo{author}{Bresson, X.}
\newblock \bibinfo{journal}{\bibinfo{title}{Graph neural networks with
  learnable structural and positional representations}}.
\newblock {\emph{\JournalTitle{CoRR}}}
  \textbf{\bibinfo{volume}{abs/2110.07875}} (\bibinfo{year}{2021}).

\bibitem{node2vec}
\bibinfo{author}{Grover, A.} \& \bibinfo{author}{Leskovec, J.}
\newblock \bibinfo{journal}{\bibinfo{title}{node2vec: Scalable feature learning
  for networks}}.
\newblock {\emph{\JournalTitle{CoRR}}}
  \textbf{\bibinfo{volume}{abs/1607.00653}} (\bibinfo{year}{2016}).

\end{thebibliography}






\appendix
\newpage

\section{Additional CDF distance metrics}

Here we show additional metrics for the CDF distances shown in Figure \ref{fig:coURL}. We see that these demonstrate a similar pattern as the mean absolute distance shown in \ref{fig:MAD}, though Mean Squared Distance demonstrates stronger clustering. \\

\begin{figure}[H]
\vspace*{-1cm}
    \centering
    \begin{subfigure}[t]{0.45\textwidth}
        \centering
        \makebox[\textwidth][c]{\includegraphics[width=1.2\textwidth]{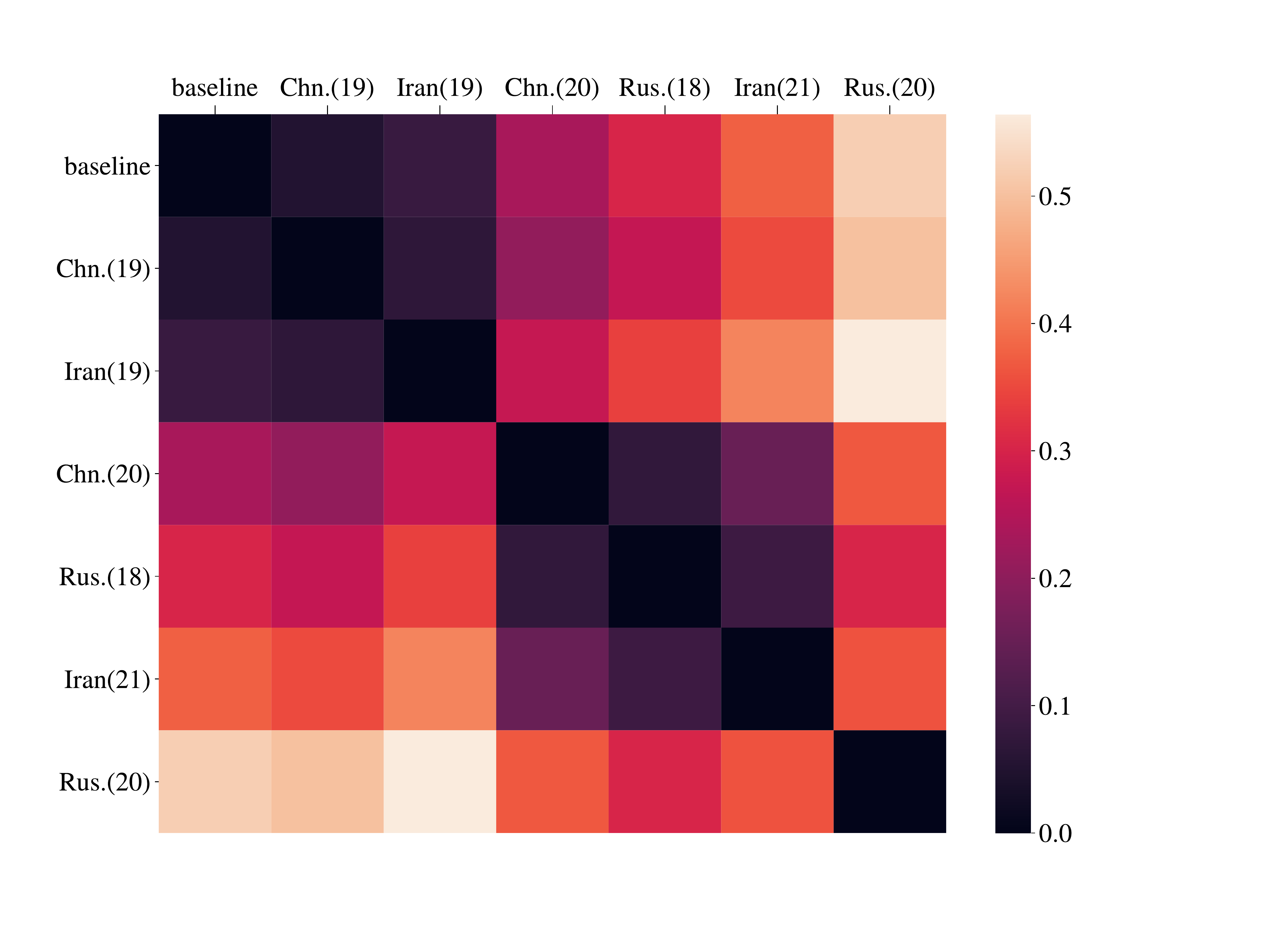}}
        \caption{ \centering Kolmogorov-Smirnov distance between each CDF in Figure \ref{fig:cdf}.} 
        \label{fig:KS}
    \end{subfigure}
    \hfill
    \begin{subfigure}[t]{0.45\textwidth}
        \centering
        \makebox[\textwidth][c]{\includegraphics[width=1.2\textwidth]{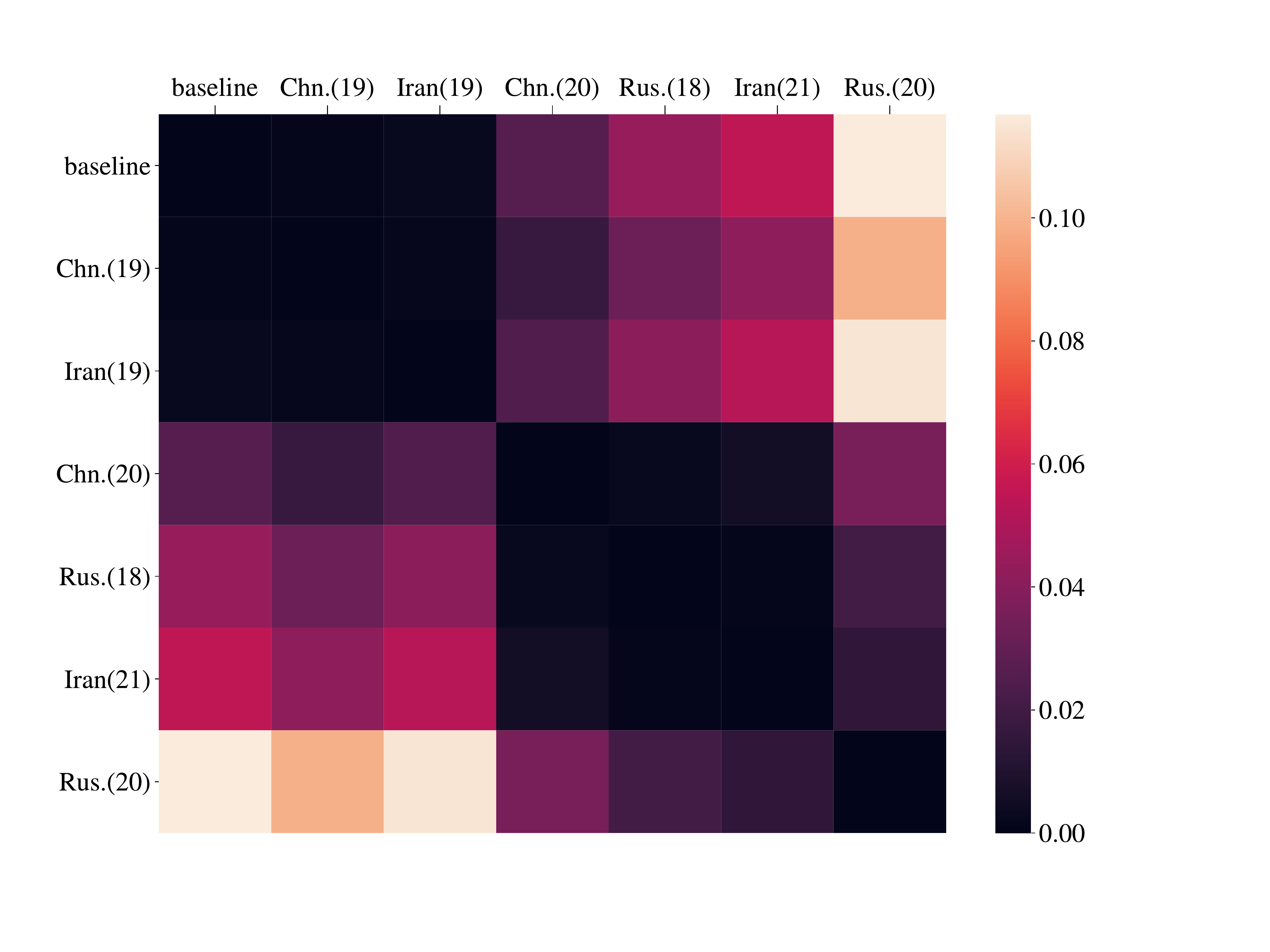}}
        \caption{\centering Mean Squared Distance between each CDF in Figure \ref{fig:cdf}.} 
        \label{fig:MSE}
    \end{subfigure}
\end{figure}

\section{Subtask F1(val/test) and AUC(test) for varying $\gamma_{\textrm{max}}$ and $k_{\textrm{top}}$}

We show here individual results for F1(val), F1(test), and AUC(test). In the top 2 panels of each figure, we show the aggregated results of Figure \ref{fig:gamma_varied}, and in the lower 6 panels we show the subtask results from which these are computed (as the harmonic mean).

\begin{figure}
    \vspace*{-1.7cm}
    \centering
    \includegraphics[width=.95\linewidth]{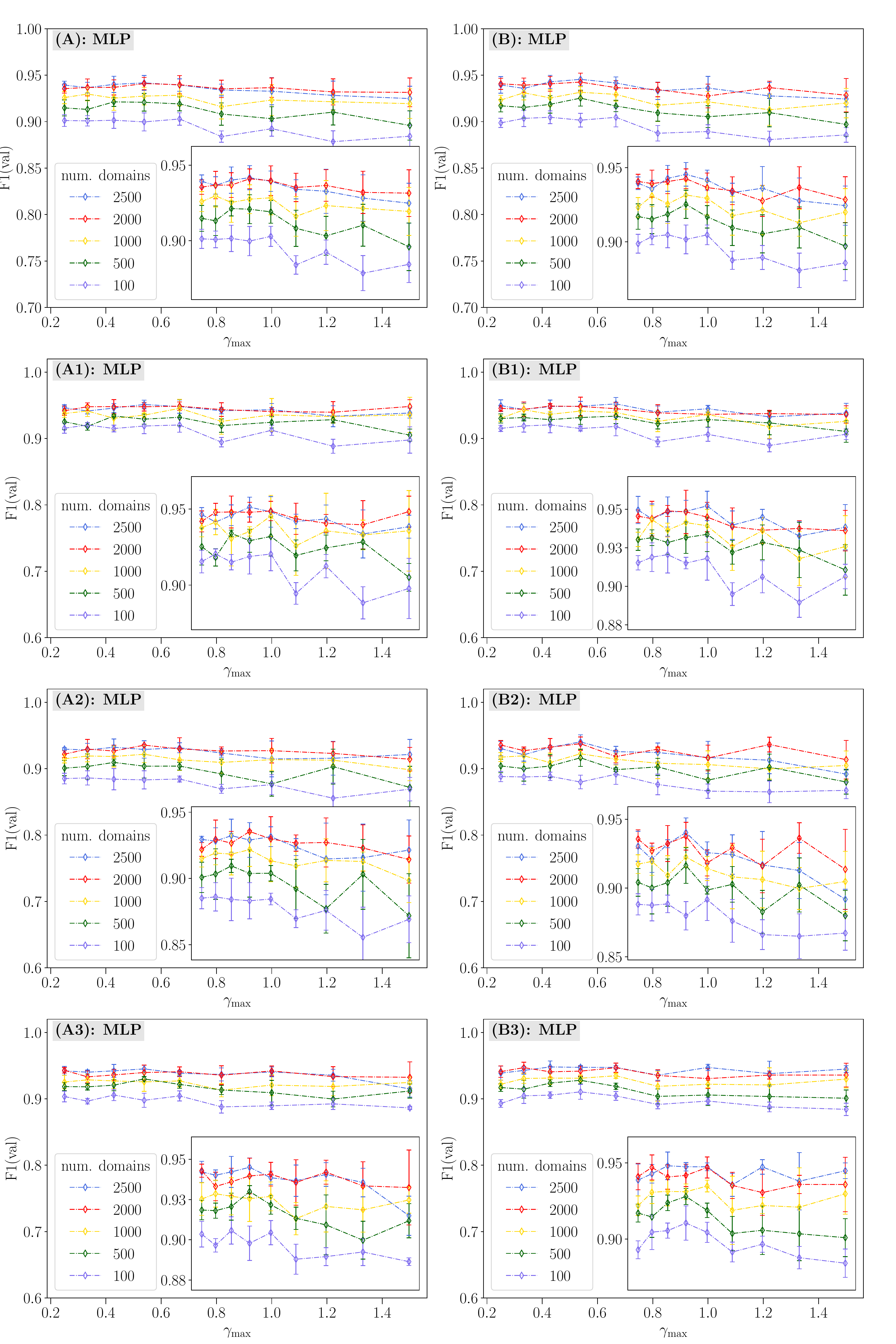}
    \caption{F1(val) for aggregated tasks (top two panels) and subtasks (bottom six panels). \underline{\textbf{Inset:}} Series replotted with a rescaled $y$-axis.}
    \label{fig:my_label1}
\end{figure}

\begin{figure}
    \vspace*{-1.7cm}
    \centering
    \includegraphics[width=.95\linewidth]{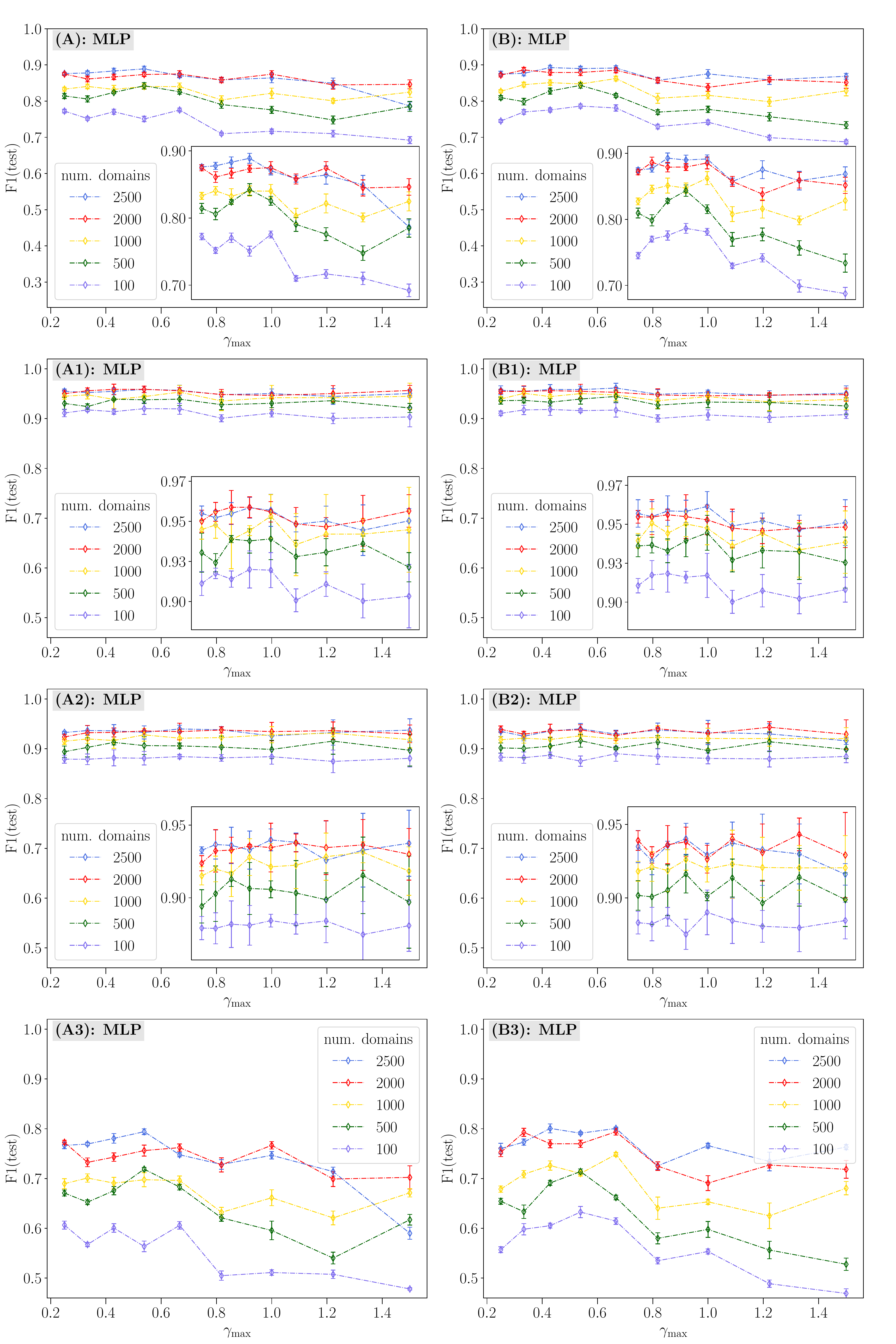}
    \caption{F1(test) for aggregated tasks (top two panels) and subtasks (bottom six panels). \underline{\textbf{Inset:}} Series replotted with a rescaled $y$-axis.}
    \label{fig:my_label2}
\end{figure}

\begin{figure}
    \vspace*{-1.7cm}
    \centering
    \includegraphics[width=.95\linewidth]{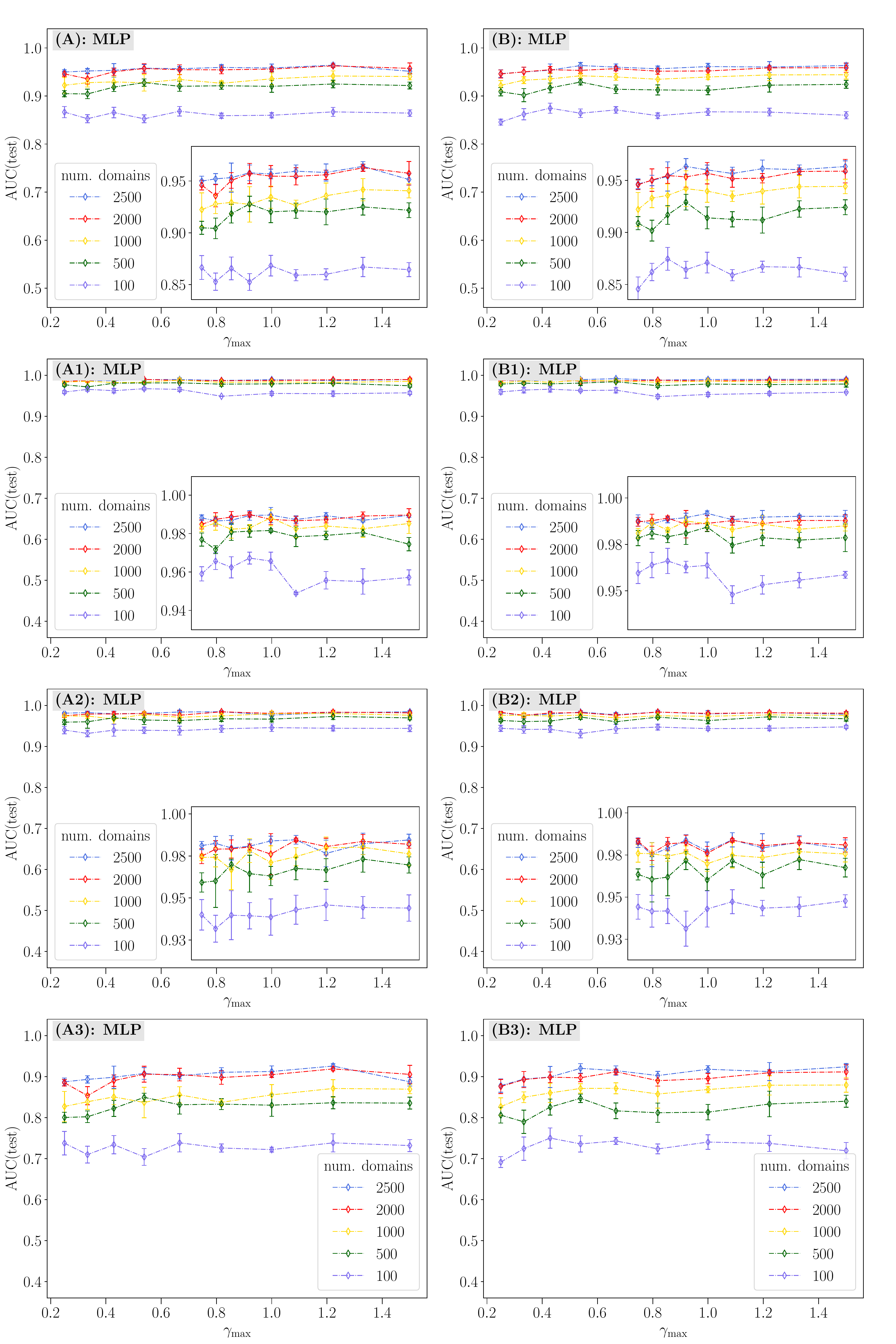}
    \caption{AUC(test) for aggregated tasks (top two panels) and subtasks (bottom six panels). \underline{\textbf{Inset:}} Series replotted with a rescaled $y$-axis.}
    \label{fig:my_label3}
\end{figure}

\begin{table}[!tb]

\newpage 

\section{Subtask level integrated gradients }

In Table \ref{tab:ig_agg}, we show the mean IG values over subtasks. Here we show explicit results for each of these subtasks. The subtask level results are largely consistent with the aggregated results, although attributions for individual features can vary by up to $\pm20$\% across subtasks. \\

\hspace{-1cm}
\begin{minipage}{.56\linewidth}
\begin{center}
\begin{tabular}{l}
      \textbf{(A1): \ Rus(18)   $ \ \rightarrow \ $   Rus(18) / Rus(20)}   \\  \hline 
\end{tabular}
\end{center}

\begin{tabular}{l|l|l|l}
feature             & IG(val.) & IG(test) & IG(base.)  \\  \hline 
\textbf{domains}    & \, $9.44 \times 10^{-2}$     &  \, $5.21 \times 10^{-2}$   &  \, $4.61 \times 10^{-1}$  \\
\textbf{node2vec}   & \, $1.54 \times 10^{-1}$     &  \, $3.40 \times 10^{-1}$   &  \, $1.98 \times 10^{-1}$  \\
\textbf{LE}         & \, $8.87 \times 10^{-5}$     &  \, $1.26 \times 10^{-4}$   &  \, $4.16 \times 10^{-5}$  \\
\textbf{RWPE}       & \, $7.44 \times 10^{-2}$     &  \, $1.77 \times 10^{-1}$   &  \, $3.23 \times 10^{-1}$  \\
\textbf{NF}         & \, $6.43 \times 10^{-1}$     &  \, $7.09 \times 10^{-1}$   &  \, $8.11 \times 10^{-1}$  \\
\,  degree          & \, $3.55 \times 10^{-1}$     &  \, $3.44 \times 10^{-1}$   &  \, $2.66 \times 10^{-1}$  \\
\,  cluster. coef.  & \, $1.70 \times 10^{-2}$       &  \, $2.56 \times 10^{-2}$   &  \, $3.34 \times 10^{-2}$  \\
\,  betweenness     & \, $1.58 \times 10^{-2}$       &  \, $2.97 \times 10^{-3}$   &  \, $7.28 \times 10^{-2}$  \\
\,  pagerank        & \, $1.32 \times 10^{-1}$      &  \, $1.78 \times 10^{-1}$   &  \, $3.44 \times 10^{-1}$  \\
\,  HITS            & \, $1.23 \times 10^{-1}$     &  \, $1.58 \times 10^{-1}$   &  \, $9.43 \times 10^{-2}$  \\
\end{tabular}
\centering

\centering
\begin{center}
\begin{tabular}{l}
      \textbf{(A2): \ Chn(19) $ \ \rightarrow \ $ Chn(19) / Chn(20)}   \\  \hline 
\end{tabular}
\end{center}

\begin{tabular}{l|l|l|l}
feature             & IG(val.) & IG(test) & IG(base.)  \\  \hline 
\textbf{domains}    & \, $8.67 \times 10^{-2}$    &  \, $1.23 \times 10^{-2}$   &  \, $4.60 \times 10^{-1}$   \\
\textbf{node2vec}   & \, $2.33 \times 10^{-1}$    &  \, $2.76 \times 10^{-1}$   &  \, $1.91 \times 10^{-1}$   \\
\textbf{LE}         & \, $8.81 \times 10^{-5}$    &  \, $2.71 \times 10^{-4}$   &  \, $3.62 \times 10^{-5}$   \\
\textbf{RWPE}       & \, $1.37 \times 10^{-1}$    &  \, $2.97 \times 10^{-1}$   &  \, $3.17 \times 10^{-1}$   \\
\textbf{NF}         & \, $7.04 \times 10^{-1}$    &  \, $7.74 \times 10^{-1}$   &  \, $8.03 \times 10^{-1}$   \\
\,  degree          & \, $3.53 \times 10^{-1}$    &  \, $2.45 \times 10^{-1}$   &  \, $2.62 \times 10^{-1}$   \\
\,  cluster. coef.  & \, $6.56 \times 10^{-3}$      &  \, $1.86 \times 10^{-2}$   &  \, $6.04 \times 10^{-3}$   \\
\,  betweenness     & \, $1.68 \times 10^{-3}$    &  \, $3.41 \times 10^{-2}$   &  \, $7.96 \times 10^{-2}$   \\
\,  pagerank        & \, $1.86 \times 10^{-1}$      &    \, $2.73 \times 10^{-1}$   &  \, $3.79 \times 10^{-1}$   \\
\,  HITS            & \, $1.57 \times 10^{-1}$    &  \, $2.04 \times 10^{-1}$   &  \, $7.74 \times 10^{-2}$   \\
\end{tabular}
\vspace*{0mm}
\centering
\centering
\begin{center}
\begin{tabular}{l}
      \textbf{(A3): \ Iran(19) $ \ \rightarrow \ $ Iran(19) / Iran(21)}   \\  \hline 
\end{tabular}
\end{center}

\begin{tabular}{l|l|l|l}
feature             & IG(val.) & IG(test) & IG(base.)  \\  \hline 
\textbf{domains}    &  \, $9.86 \times 10^{-2}$   &  \, $2.55 \times 10^{-1}$   & \, $4.34 \times 10^{-1}$   \\
\textbf{node2vec}   &  \, $2.47 \times 10^{-1}$   &  \, $4.10 \times 10^{-1}$   & \, $1.71 \times 10^{-1}$   \\
\textbf{LE}         &  \, $1.09 \times 10^{-4}$   &  \, $1.20 \times 10^{-3}$   & \, $3.98 \times 10^{-5}$   \\
\textbf{RWPE}       &  \, $1.44 \times 10^{-1}$   &  \, $3.76 \times 10^{-1}$   & \, $3.00 \times 10^{-1}$   \\
\textbf{NF}         &  \, $7.54 \times 10^{-1}$   &  \, $1.12 \times 10^{0}$    & \, $7.57 \times 10^{-1}$   \\
\,  degree          &  \, $3.79 \times 10^{-1}$   &  \, $3.18 \times 10^{-1}$   & \, $2.52 \times 10^{-1}$   \\
\,  cluster. coef.  &  \, $4.97 \times 10^{-3}$     &  \, $2.32 \times 10^{-3}$   & \, $5.35 \times 10^{-3}$   \\
\,  betweenness     &  \, $9.96 \times 10^{-3}$   &  \, $9.00 \times 10^{-2}$   & \, $8.25 \times 10^{-2}$   \\
\,  pagerank        &  \, $2.06 \times 10^{-1}$     &  \, $4.09 \times 10^{-1}$     & \, $3.57 \times 10^{-1}$     \\
\,  HITS            &  \, $1.55 \times 10^{-1}$   &  \, $2.97 \times 10^{-1}$   & \, $6.06 \times 10^{-2}$   \\
\end{tabular}
    \end{minipage}%
    \begin{minipage}{.56\linewidth}
\begin{center}
\begin{tabular}{l}
      \textbf{(B1): \ Chn(19) + Iran(19)  $\rightarrow$ Rus(18) / Rus(20)}   \\  \hline 
\end{tabular}
\end{center}

\begin{tabular}{l|l|l|l}
feature             & IG(val.) & IG(test) & IG(base.)  \\  \hline 
\textbf{domains}    & \, $9.79 \times 10^{-2}$     &  \, $4.84 \times 10^{-2}$   &  \, $4.78 \times 10^{-1}$   \\
\textbf{node2vec}   & \, $1.51 \times 10^{-1}$     &  \, $3.39 \times 10^{-1}$   &  \, $2.09 \times 10^{-1}$   \\
\textbf{LE}         & \, $1.04 \times 10^{-4}$     &  \, $1.31 \times 10^{-4}$   &  \, $4.93 \times 10^{-5}$   \\
\textbf{RWPE}       & \, $7.26 \times 10^{-2}$     &  \, $1.74 \times 10^{-1}$   &  \, $3.44 \times 10^{-1}$   \\
\textbf{NF}         & \, $5.85 \times 10^{-1}$     &  \, $6.40 \times 10^{-1}$   &  \, $8.18 \times 10^{-1}$   \\
\,  degree          & \, $3.54 \times 10^{-1}$     &  \, $3.45 \times 10^{-1}$   &  \, $2.81 \times 10^{-1}$   \\
\,  cluster. coef.  & \, $4.64 \times 10^{-3}$       &  \, $3.48 \times 10^{-3}$   &    \, $9.27 \times 10^{-3}$ \\
\,  betweenness     & \, $9.28 \times 10^{-3}$       &  \, $5.13 \times 10^{-3}$   &    \, $8.90 \times 10^{-2}$ \\
\,  pagerank        & \, $1.19 \times 10^{-1}$       &  \, $1.66 \times 10^{-1}$   &    \, $3.70 \times 10^{-1}$ \\
\,  HITS            & \, $9.81 \times 10^{-2}$     &  \, $1.21 \times 10^{-1}$   &  \, $6.91 \times 10^{-2}$   \\
\end{tabular}
\centering

\centering
\begin{center}
\begin{tabular}{l}
      \textbf{(B2): \ Rus(18) + Iran(19)  $\rightarrow$ Chn(19) / Chn(20)}   \\  \hline 
\end{tabular}
\end{center}

\begin{tabular}{l|l|l|l}
feature             & IG(val.) & IG(test) & IG(base.)  \\  \hline 
\textbf{domains}    &  \, $5.05 \times 10^{-2}$   & \, $1.38 \times 10^{-2}$    & \, $4.21 \times 10^{-1}$   \\
\textbf{node2vec}   &  \, $2.06 \times 10^{-1}$   & \, $3.13 \times 10^{-1}$    & \, $1.73 \times 10^{-1}$   \\
\textbf{LE}         &  \, $9.03 \times 10^{-5}$   & \, $3.42 \times 10^{-4}$    & \, $4.14 \times 10^{-5}$   \\
\textbf{RWPE}       &  \, $1.18 \times 10^{-1}$   & \, $3.05 \times 10^{-1}$    & \, $3.03 \times 10^{-1}$   \\
\textbf{NF}         &  \, $6.11 \times 10^{-1}$   & \, $8.58 \times 10^{-1}$    & \, $7.51 \times 10^{-1}$   \\
\,  degree          &  \, $3.05 \times 10^{-1}$   & \, $2.78 \times 10^{-1}$    & \, $2.54 \times 10^{-1}$   \\
\,  cluster. coef.  &  \, $5.05 \times 10^{-3}$   & \, $1.36 \times 10^{-2}$    & \, $7.76 \times 10^{-3}$   \\
\,  betweenness     &  \, $9.21 \times 10^{-3}$   & \, $4.90 \times 10^{-2}$    & \, $7.45 \times 10^{-2}$   \\
\,  pagerank        &  \, $1.65 \times 10^{-1}$   & \, $3.09 \times 10^{-1}$    & \, $3.53 \times 10^{-1}$   \\
\,  HITS            &  \, $1.27 \times 10^{-1}$   & \, $2.09 \times 10^{-1}$    & \, $6.25 \times 10^{-2}$   \\
\end{tabular}

\vspace*{0mm}
\centering
\centering
\begin{center}
\begin{tabular}{l}
      \textbf{(B3): \ Chn(19) + Rus(18)  $\rightarrow$ Iran(19) / Iran(21)}   \\  \hline 
\end{tabular}
\end{center}

\begin{tabular}{l|l|l|l}
feature             & IG(val.) & IG(test) & IG(base.)  \\  \hline 
\textbf{domains}    & \, $8.67 \times 10^{-2}$    & \, $2.17 \times 10^{-1}$    & \, $4.51 \times 10^{-1}$   \\
\textbf{node2vec}   & \, $2.25 \times 10^{-1}$    & \, $3.61 \times 10^{-1}$    & \, $1.97 \times 10^{-1}$   \\
\textbf{LE}         & \, $1.05 \times 10^{-4}$    & \, $1.08 \times 10^{-3}$    & \, $4.88 \times 10^{-5}$   \\
\textbf{RWPE}       & \, $1.55 \times 10^{-1}$    & \, $3.83 \times 10^{-1}$    & \, $3.33 \times 10^{-1}$   \\
\textbf{NF}         & \, $7.18 \times 10^{-1}$    & \, $1.01 \times 10^{0}$     & \, $7.83 \times 10^{-1}$   \\
\,  degree          & \, $3.33 \times 10^{-1}$    & \, $2.69 \times 10^{-1}$    & \, $2.60 \times 10^{-1}$   \\
\,  cluster. coef.  & \, $5.06 \times 10^{-3}$      & \, $6.10 \times 10^{-4}$      & \, $4.63 \times 10^{-3}$   \\
\,  betweenness     & \, $7.53 \times 10^{-3}$    & \, $6.51 \times 10^{-2}$    & \, $7.88 \times 10^{-2}$   \\
\,  pagerank        & \, $1.91 \times 10^{-1}$      & \, $3.47 \times 10^{-1}$      & \, $3.58 \times 10^{-1}$   \\
\,  HITS            & \, $1.81 \times 10^{-1}$    & \, $3.32 \times 10^{-1}$    & \, $8.14 \times 10^{-2}$   \\
\end{tabular}
    \end{minipage} 
\caption{IG values of individual subtasks.}
\label{tab:ig_trials}
\end{table}

\begin{table}[]
\nopagebreak
\vspace*{-1.2cm}
\section{Integrated gradients by domain}
Here we show the most significant domain level integrated gradients. We leave these values signed since they are largely consistent across data splits and subtasks. The signs do not necessarily indicate with certainty whether a feature was used to make a positive ($y=1$) or negative ($y=0$) prediction, though one expects a reasonable correspondence.

\vspace{2mm}
      \centering
\begin{center}
\begin{tabular}{l}
      \textbf{(A1): \ Rus(18)   $ \ \rightarrow \ $   Rus(18) / Rus(20)}   \\  \hline 
\end{tabular}
\end{center}
\vspace*{-1mm}

\begin{tabular}{l|l}
domain              & IG(val.)   \\  \hline 
untappd.com &  \,$4.65 \times 10^{-3}$ \\
senate.gov &  \,$3.80 \times 10^{-3}$ \\
nbcnews.com &  \,$2.35 \times 10^{-3}$ \\
tumblr.com &  -$2.33 \times 10^{-3}$ \\
pscp.tv &  \,$1.85 \times 10^{-3}$ \\
washingtonexam[*].com &  \,$1.82 \times 10^{-3}$ \\
foursquare.com &  -$1.77 \times 10^{-3}$ \\
furnishhome.ru &  -$1.71 \times 10^{-3}$ \\
house.gov &  -$1.65 \times 10^{-3}$ \\
wordpress.com &  -$1.55 \times 10^{-3}$ \\
washingtonpost.com &  -$1.53 \times 10^{-3}$ \\
yahoo.com &  -$1.53 \times 10^{-3}$ \\
fllwrs.com &  -$1.34 \times 10^{-3}$ \\
twittascope.com &  \,$1.32 \times 10^{-3}$ \\
rawstory.com &  -$1.28 \times 10^{-3}$ \\  

\end{tabular}
\begin{tabular}{l|l}
domain              & IG(test)   \\  \hline 
inc.com &  -$8.51 \times 10^{-3}$ \\
soundcloud.com &  \,$5.06 \times 10^{-3}$ \\
yahoo.com &  -$2.82 \times 10^{-3}$ \\
nbcnews.com &  \,$1.81 \times 10^{-3}$ \\
senate.gov &  \,$1.60 \times 10^{-3}$ \\
espn.com &  -$1.58 \times 10^{-3}$ \\
techchrunch.com &  -$1.52 \times 10^{-3}$ \\
change.org &  \,$1.47 \times 10^{-3}$ \\
ebay.com &  -$1.39 \times 10^{-3}$ \\
foursquare.com &  -$1.21 \times 10^{-3}$ \\
usatoday.com &  -$1.18 \times 10^{-3}$ \\
nationalreview.com &  \,$9.91 \times 10^{-4}$ \\
washingtonpost.com &  -$9.65 \times 10^{-4}$ \\
smotrim.ru &  -$9.56 \times 10^{-4}$ \\
sba.gov &  \,$9.19 \times 10^{-4}$ \\

\end{tabular}
\begin{tabular}{l|l}
domain              & IG(base.)   \\  \hline 
nytimes.com &  -$2.08 \times 10^{-2}$ \\
abouthub.info &  -$1.30 \times 10^{-2}$ \\
conta.cc &  -$1.08 \times 10^{-2}$ \\
politico.com &  -$1.08 \times 10^{-2}$ \\
senate.gov &  \,$9.19 \times 10^{-3}$ \\
abcnews.com &  -$6.90 \times 10^{-3}$ \\
reddit.com &  -$6.53 \times 10^{-3}$ \\
github.com &  -$6.22 \times 10^{-3}$ \\
mailchimp.com &  -$5.98 \times 10^{-3}$ \\
wsj.com &  -$5.95 \times 10^{-3}$ \\
wabcradio.com &  -$5.86 \times 10^{-3}$ \\
instagram.com &  -$5.78 \times 10^{-3}$ \\
hubs.ly &  -$5.59 \times 10^{-3}$ \\
foxnews.com &  -$5.54 \times 10^{-3}$ \\
msn.com &  -$5.39 \times 10^{-3}$ \\

\end{tabular}
\centering

\centering
\begin{center}
\begin{tabular}{l}
      \textbf{(A2): \ Chn(19) $ \ \rightarrow \ $ Chn(19) / Chn(20)}   \\  \hline 
\end{tabular}
\end{center}
\vspace*{-1mm}

\begin{tabular}{l|l}
domain              & IG(val.)   \\  \hline 
cnn.com &  -$2.88 \times 10^{-3}$ \\
washingtonpost.com &  -$2.49 \times 10^{-3}$ \\
cdt.org &  \, $2.16 \times 10^{-3}$ \\
wordpress.com &  -$1.63 \times 10^{-3}$ \\
wsj.com &  -$1.59 \times 10^{-3}$ \\
medium.com &  -$1.11 \times 10^{-3}$ \\
bloomberg.com &  \, $9.32 \times 10^{-4}$ \\
eventbrite.com &  -$8.78 \times 10^{-4}$ \\
theconversation.com &  \, $7.66 \times 10^{-4}$ \\
yahoo.com &  -$7.61 \times 10^{-4}$ \\
ap.org &  \, $7.39 \times 10^{-4}$ \\
inewsource.org &  \, $6.83 \times 10^{-4}$ \\
www.gov.uk &  -$6.53 \times 10^{-4}$ \\
mktw.net &  -$6.42 \times 10^{-4}$ \\
sen.gov &  -$6.28 \times 10^{-4}$ \\

\end{tabular}
\begin{tabular}{l|l}
domain              & IG(test)   \\  \hline 
justunfollow.com &  \, $5.48 \times 10^{-3}$ \\
etsy.com &  -$7.58 \times 10^{-4}$ \\
tumblr.com &  -$5.40 \times 10^{-4}$ \\
yahoo.com &  -$4.56 \times 10^{-4}$ \\
cnn.com &  -$4.53 \times 10^{-4}$ \\
foursquare.com &  -$3.83 \times 10^{-4}$ \\
house.gov &  -$3.57 \times 10^{-4}$ \\
medium.com &  -$3.50 \times 10^{-4}$ \\
apnews.com &  \, $2.57 \times 10^{-4}$ \\
conservativejunc[*].com &  \, $2.54 \times 10^{-4}$ \\
twittascope.com &  \, $2.25 \times 10^{-4}$ \\
sba.gov &  \, $2.12 \times 10^{-4}$ \\
theconversation.com &  -$1.62 \times 10^{-4}$ \\
wired.com &  -$1.36 \times 10^{-4}$ \\
github.com &  -$1.29 \times 10^{-4}$ \\

\end{tabular}
\begin{tabular}{l|l}
domain              & IG(base.)   \\  \hline 
nytimes.com &  -$2.08 \times 10^{-2}$ \\
abouthub.info &  -$1.30 \times 10^{-2}$ \\
conta.cc &  -$1.08 \times 10^{-2}$ \\
politico.com &  -$1.08 \times 10^{-2}$ \\
senate.gov &  \,$9.19 \times 10^{-3}$ \\
abcnews.com &  -$6.90 \times 10^{-3}$ \\
reddit.com &  -$6.53 \times 10^{-3}$ \\
github.com &  -$6.22 \times 10^{-3}$ \\
mailchimp.com &  -$5.98 \times 10^{-3}$ \\
wsj.com &  -$5.95 \times 10^{-3}$ \\
wabcradio.com &  -$5.86 \times 10^{-3}$ \\
instagram.com &  -$5.78 \times 10^{-3}$ \\
hubs.ly &  -$5.59 \times 10^{-3}$ \\
foxnews.com &  -$5.54 \times 10^{-3}$ \\
msn.com &  -$5.39 \times 10^{-3}$ \\

\end{tabular}
\vspace*{0mm}
\centering
\centering
\begin{center}
\begin{tabular}{l}
      \textbf{(A3): \ Iran(19) $ \ \rightarrow \ $ Iran(19) / Iran(21)}   \\  \hline 
\end{tabular}
\end{center}
\vspace*{-1mm}

\begin{tabular}{l|l}
domain              & IG(val.)   \\  \hline 
huffpost.com &  -$4.25 \times 10^{-3}$ \\
foursquare.com &  -$3.89 \times 10^{-3}$ \\
medium.com &  -$3.68 \times 10^{-3}$ \\
untappd.com &  \, $3.58 \times 10^{-3}$ \\
cnn.com &  -$3.09 \times 10^{-3}$ \\
senate.gov &  \, $2.82 \times 10^{-3}$ \\
house.gov &  -$2.66 \times 10^{-3}$ \\
etsy.com &  -$2.41 \times 10^{-3}$ \\
apple.news &  -$2.34 \times 10^{-3}$ \\
economist.com &  \, $2.19 \times 10^{-3}$ \\
abouthub.info &  -$2.14 \times 10^{-3}$ \\
yahoo.com &  -$2.13 \times 10^{-3}$ \\
amazon.com &  -$2.05 \times 10^{-3}$ \\
twittascope.com &  -$1.79 \times 10^{-3}$ \\
tumblr.com &  -$1.73 \times 10^{-3}$ \\

\end{tabular}
\begin{tabular}{l|l}
domain              & IG(test)   \\  \hline 
untappd.com &  \, $2.41 \times 10^{-2}$ \\
gofundme.com &  \, $1.87 \times 10^{-2}$ \\
twittascope.com &  -$1.34 \times 10^{-2}$ \\
bbc.co.uk &  \, $1.24 \times 10^{-2}$ \\
wired.com &  -$9.16 \times 10^{-3}$ \\
iheart.com &  \, $8.79 \times 10^{-3}$ \\
npr.org &  \, $8.12 \times 10^{-3}$ \\
senate.gov &  \, $7.63 \times 10^{-3}$ \\
washingtonexaminer.com &  \, $6.55 \times 10^{-3}$ \\
house.gov &  -$6.23 \times 10^{-3}$ \\
justunfollow.com &  \, $6.04 \times 10^{-3}$ \\
wordpress.com &  -$5.87 \times 10^{-3}$ \\
tumblr.com &  -$5.75 \times 10^{-3}$ \\
economist.com &  \, $5.54 \times 10^{-3}$ \\
bloomberg.com &  \, $5.01 \times 10^{-3}$ \\

\end{tabular}
\begin{tabular}{l|l}
domain              & IG(base.)   \\  \hline 
nytimes.com &  -$1.83 \times 10^{-2}$ \\
abouthub.info &  -$1.12 \times 10^{-2}$ \\
politico.com &  -$1.00 \times 10^{-2}$ \\
conta.cc &  -$9.88 \times 10^{-3}$ \\
senate.gov &  \, $9.27 \times 10^{-3}$ \\
abcnews.com &  -$6.67 \times 10^{-3}$ \\
reddit.com &  -$5.96 \times 10^{-3}$ \\
wsj.com &  -$5.94 \times 10^{-3}$ \\
github.com &  -$5.58 \times 10^{-3}$ \\
mailchimp.com &  -$5.53 \times 10^{-3}$ \\
wabcradio.com &  -$5.32 \times 10^{-3}$ \\
hubs.ly &  -$5.23 \times 10^{-3}$ \\
instagram.com &  -$5.16 \times 10^{-3}$ \\
foxnews.com &  -$5.10 \times 10^{-3}$ \\
msn.com &  -$4.86 \times 10^{-3}$ \\

\end{tabular}
\end{table}
\vspace*{-2cm}
\begin{table}[]
\begin{center}
\begin{tabular}{l}
      \textbf{(B1): \ Chn(19) + Iran(19)  $\rightarrow$ Rus(18) / Rus(20)}   \\  \hline 
\end{tabular}
\end{center}
\vspace*{-1mm}

\begin{tabular}{l|l}
domain              & IG(val.)   \\  \hline 
nbcnews.com &  \, $3.28 \times 10^{-3}$ \\
tumblr.com &  -$2.85 \times 10^{-3}$ \\
house.gov &  -$2.78 \times 10^{-3}$ \\
senate.gov &  \, $2.50 \times 10^{-3}$ \\
pscp.tv &  \, $2.41 \times 10^{-3}$ \\
foursquare.com &  -$2.16 \times 10^{-3}$ \\
wordpress.com &  -$2.05 \times 10^{-3}$ \\
thehill.com &  -$2.01 \times 10^{-3}$ \\
washingtonpost.com &  -$1.95 \times 10^{-3}$ \\
furnishhome.ru &  -$1.84 \times 10^{-3}$ \\
bbc.co.uk &  \, $1.51 \times 10^{-3}$ \\
forbes.com &  -$1.44 \times 10^{-3}$ \\
yahoo.com &  -$1.41 \times 10^{-3}$ \\
bglradio.net &  -$1.41 \times 10^{-3}$ \\
fllwrs.com &  -$1.36 \times 10^{-3}$ \\

\end{tabular}
\begin{tabular}{l|l}
domain              & IG(test)   \\  \hline 
inc.com &  -$7.94 \times 10^{-3}$ \\
yahoo.com &  -$3.76 \times 10^{-3}$ \\
nbcnews.com &  \, $2.32 \times 10^{-3}$ \\
foursquare.com &  -$1.88 \times 10^{-3}$ \\
soundcloud.com &  \, $1.69 \times 10^{-3}$ \\
espn.com &  -$1.34 \times 10^{-3}$ \\
usatoday.com &  -$1.16 \times 10^{-3}$ \\
washingtonpost.com &  -$1.15 \times 10^{-3}$ \\
change.org &  \, $1.11 \times 10^{-3}$ \\
apnews.com &  \, $1.06 \times 10^{-3}$ \\
house.gov &  -$1.05 \times 10^{-3}$ \\
sba.gov &  \, $9.44 \times 10^{-4}$ \\
senate.gov &  \, $9.17 \times 10^{-4}$ \\
pinterest.com &  -$8.24 \times 10^{-4}$ \\
huffpost.com &  -$7.92 \times 10^{-4}$ \\

\end{tabular}
\begin{tabular}{l|l}
domain              & IG(base.)   \\  \hline 
nytimes.com &  -$2.07 \times 10^{-2}$ \\
abouthub.info &  -$1.22 \times 10^{-2}$ \\
politico.com &  -$1.09 \times 10^{-2}$ \\
conta.cc &  -$1.08 \times 10^{-2}$ \\
senate.gov &  \, $8.49 \times 10^{-3}$ \\
abcnews.com &  -$7.85 \times 10^{-3}$ \\
reddit.com &  -$6.85 \times 10^{-3}$ \\
mailchimp.com &  -$6.82 \times 10^{-3}$ \\
wsj.com &  -$6.40 \times 10^{-3}$ \\
github.com &  -$6.19 \times 10^{-3}$ \\
wabcradio.com &  -$6.12 \times 10^{-3}$ \\
hubs.ly &  -$5.82 \times 10^{-3}$ \\
foxnews.com &  -$5.82 \times 10^{-3}$ \\
instagram.com &  -$5.75 \times 10^{-3}$ \\
msn.com &  -$5.26 \times 10^{-3}$ \\

\end{tabular}
\centering

\centering
\begin{center}
\begin{tabular}{l}
      \textbf{(B2): \ Rus(18) + Iran(19)  $\rightarrow$ Chn(19) / Chn(20)}   \\  \hline 
\end{tabular}
\end{center}
\vspace*{-1mm}

\begin{tabular}{l|l}
domain              & IG(val.)   \\  \hline 
washingtonpost.com &  -$2.20 \times 10^{-3}$ \\
cnn.com &  -$2.02 \times 10^{-3}$ \\
twittascope.com &  -$1.56 \times 10^{-3}$ \\
wordpress.com &  -$1.09 \times 10^{-3}$ \\
senate.gov &  \, $1.04 \times 10^{-3}$ \\
cdt.org &  \, $9.19 \times 10^{-4}$ \\
yahoo.com &  -$8.57 \times 10^{-4}$ \\
sen.gov &  -$8.55 \times 10^{-4}$ \\
theconversation.com &  \, $8.21 \times 10^{-4}$ \\
medium.com &  -$7.94 \times 10^{-4}$ \\
epsomguardian.co.uk &  \, $7.85 \times 10^{-4}$ \\
msn.com &  -$7.39 \times 10^{-4}$ \\
bloomberg.com &  \, $7.12 \times 10^{-4}$ \\
furnishhome.ru &  -$6.89 \times 10^{-4}$ \\
inewsource.org &  \, $6.69 \times 10^{-4}$ \\ 

\end{tabular}
\begin{tabular}{l|l}
domain              & IG(test)   \\  \hline 
justunfollow.com &  \, $6.73 \times 10^{-3}$ \\
tumblr.com &  -$6.54 \times 10^{-4}$ \\
etsy.com &  -$5.16 \times 10^{-4}$ \\
yahoo.com &  -$5.03 \times 10^{-4}$ \\
cnn.com &  -$4.89 \times 10^{-4}$ \\
foursquare.com &  -$4.69 \times 10^{-4}$ \\
house.gov &  -$3.58 \times 10^{-4}$ \\
conservativejun[*].com &  \, $3.32 \times 10^{-4}$ \\
medium.com &  -$2.83 \times 10^{-4}$ \\
theconversation.com &  -$2.27 \times 10^{-4}$ \\
apnews.com &  \, $2.06 \times 10^{-4}$ \\
github.com &  -$1.98 \times 10^{-4}$ \\
wired.com &  -$1.94 \times 10^{-4}$ \\
sba.gov &  \, $1.66 \times 10^{-4}$ \\
cnsnews.com &  -$1.49 \times 10^{-4}$ \\

\end{tabular}
\begin{tabular}{l|l}
domain              & IG(base.)   \\  \hline 
nytimes.com &  -$1.85 \times 10^{-2}$ \\
abouthub.info &  -$1.06 \times 10^{-2}$ \\
conta.cc &  -$1.04 \times 10^{-2}$ \\
politico.com &  -$1.00 \times 10^{-2}$ \\
senate.gov &  \, $8.92 \times 10^{-3}$ \\
abcnews.com &  -$7.24 \times 10^{-3}$ \\
reddit.com &  -$6.16 \times 10^{-3}$ \\
wsj.com &  -$5.92 \times 10^{-3}$ \\
mailchimp.com &  -$5.55 \times 10^{-3}$ \\
wabcradio.com &  -$5.50 \times 10^{-3}$ \\
hubs.ly &  -$5.48 \times 10^{-3}$ \\
instagram.com &  -$5.29 \times 10^{-3}$ \\
foxnews.com &  -$5.28 \times 10^{-3}$ \\
msn.com &  -$5.02 \times 10^{-3}$ \\
github.com &  -$4.80 \times 10^{-3}$ \\

\end{tabular}

\vspace*{0mm}
\centering
\centering
\begin{center}
\begin{tabular}{l}
      \textbf{(B3): \ Chn(19) + Rus(18)  $\rightarrow$ Iran(19) / Iran(21)}   \\  \hline 
\end{tabular}
\end{center}
\vspace*{-1mm}

\begin{tabular}{l|l}
domain              & IG(val.)   \\  \hline 
foursquare.com &  -$4.15 \times 10^{-3}$ \\
cnn.com &  -$3.91 \times 10^{-3}$ \\
house.gov &  -$3.85 \times 10^{-3}$ \\
medium.com &  -$3.66 \times 10^{-3}$ \\
huffpost.com &  -$3.38 \times 10^{-3}$ \\
senate.gov &  \, $2.67 \times 10^{-3}$ \\
apple.news &  -$2.44 \times 10^{-3}$ \\
amazon.com &  -$2.10 \times 10^{-3}$ \\
greenlifestylech[*].com &  \, $1.84 \times 10^{-3}$ \\
tumblr.com &  -$1.70 \times 10^{-3}$ \\
usa.gov &  -$1.63 \times 10^{-3}$ \\
economist.com &  \, $1.62 \times 10^{-3}$ \\
theguardian.com &  -$1.51 \times 10^{-3}$ \\
yahoo.com &  -$1.46 \times 10^{-3}$ \\
conta.cc &  -$1.37 \times 10^{-3}$ \\

\end{tabular}
\begin{tabular}{l|l}
domain              & IG(test)   \\  \hline 
gofundme.com &  \, $1.80 \times 10^{-2}$ \\
bbc.co.uk &  \, $1.28 \times 10^{-2}$ \\
twittascope.com &  -$9.93 \times 10^{-3}$ \\
house.gov &  -$9.60 \times 10^{-3}$ \\
iheart.com &  \, $7.84 \times 10^{-3}$ \\
wordpress.com &  -$7.82 \times 10^{-3}$ \\
npr.org &  \, $7.49 \times 10^{-3}$ \\
cnn.com &  -$6.56 \times 10^{-3}$ \\
washingtonex[*].com &  \, $6.13 \times 10^{-3}$ \\
senate.gov &  \, $6.10 \times 10^{-3}$ \\
justunfollow.com &  \, $5.54 \times 10^{-3}$ \\
bloomberg.com &  \, $5.52 \times 10^{-3}$ \\
tumblr.com &  -$5.47 \times 10^{-3}$ \\
wired.com &  -$5.17 \times 10^{-3}$ \\
economist.com &  \, $4.89 \times 10^{-3}$ \\

\end{tabular}
\begin{tabular}{l|l}
domain              & IG(base.)   \\  \hline 
nytimes.com &  -$1.95 \times 10^{-2}$ \\
conta.cc &  -$1.12 \times 10^{-2}$ \\
abouthub.info &  -$1.09 \times 10^{-2}$ \\
politico.com &  -$1.04 \times 10^{-2}$ \\
senate.gov &  \, $9.25 \times 10^{-3}$ \\
abcnews.com &  -$7.66 \times 10^{-3}$ \\
reddit.com &  -$6.26 \times 10^{-3}$ \\
wsj.com &  -$5.96 \times 10^{-3}$ \\
github.com &  -$5.93 \times 10^{-3}$ \\
mailchimp.com &  -$5.86 \times 10^{-3}$ \\
wabcradio.com &  -$5.82 \times 10^{-3}$ \\
instagram.com &  -$5.67 \times 10^{-3}$ \\
foxnews.com &  -$5.65 \times 10^{-3}$ \\
hubs.ly &  -$5.40 \times 10^{-3}$ \\
msn.com &  -$5.10 \times 10^{-3}$ \\

\end{tabular}

\caption{IG values for individual domains in each trial and val, test, and baseline set. For long domain names, [*] denotes truncation.}
\label{table:ig_domains}
\end{table}
 
\vspace*{-1cm}
\newpage
\begin{table}
\vspace*{-0.8cm}
\section{Frequent and Removed domains}

Here we show the most frequent domains which are retained or censored in each dataset. We specifically show the result at the most stringent threshold $\gamma_{\textrm{max}}=0.4$ where MLP still demonstrates strong performance.

\caption*{\huge{\textbf{\underline{Baseline}}}}
    \begin{minipage}{.33\linewidth}
      \centering
\begin{tabular}{l|r}
             \textbf{indicator} & counts \\ \hline
UNCOMMON &             770187 \\ 
instagram.com &         283505 \\ 
twittascope.com &        91120 \\ 
house.gov &              68485 \\ 
nytimes.com &            58666 \\ 
washingtonpost.com &     37356 \\ 
senate.gov &             28874 \\ 
usa.gov &                20972 \\ 
theguardian.com &        20216 \\ 
huffpost.com &           19937 \\ 
wsj.com &                19904 \\ 
foxnews.com &            19632 \\ 
tumblr.com &             19555 \\ 
pinterest.com &          19031 \\ 
bloomberg.com &          18553 \\ 
foursquare.com &         18350 \\ 
wordpress.com &          17313 \\ 
soundcloud.com &         17177 \\ 
fllwrs.com &             16862 \\ 
apnews.com &             16724 \\ 
swarmapp.com &           15441 \\ 
amazon.com &             14282 \\ 
cnn.com &                14275 \\ 
pscp.tv &                13428 \\ 
usatoday.com &           12696 \\ 
politico.com &           12322 \\ 
thehill.com &            12002 \\ 
medium.com &             11073 \\ 
etsy.com &                9797 \\ 
bbc.co.uk &               9624 \\ 
npr.org &                 9299 \\ 
forbes.com &              8892 \\ 
cnbc.com &                8800 \\ 
linkedin.com &            8664 \\ 
apple.com &               8524 \\ 
github.com &              8477 \\ 
ebay.com &                8190 \\ 
eventbrite.com &          8082 \\ 
crowdfireapp.com &        8039 \\ 
untappd.com &             8009 \\ 
iacr.org &                7715 \\ 
yahoo.com &               6787 \\ 
latimes.com &             6756 \\ 
conta.cc&                6554 \\ 
cbsnews.com &             6546 \\ 
buzzfeed.com &            6486 \\ 
apple.news &              6427 \\ 
www.gov.uk &              6349 \\ 
time.com &                5862 \\ 
abouthub.info &           5761 \\ 
vox.com &                 5596 \\ 
flipit.com &              5560 \\  
        \end{tabular}
    \end{minipage}%
    \begin{minipage}{.33\linewidth}
      \centering
        \begin{tabular}{l|r}
            \textbf{indicator}  & tf-idf \\ \hline
instagram.com &         1464.24 \\ 
house.gov &               511.7 \\ 
fllwrs.com &             477.48 \\ 
soundcloud.com &         417.33 \\ 
twittascope.com &        415.91 \\ 
nytimes.com &            380.88 \\ 
foxnews.com &            289.61 \\ 
espn.com &               288.08 \\ 
washingtonpost.com &     286.39 \\ 
pscp.tv &                 281.8 \\ 
github.com &             250.51 \\ 
www.gov.uk &             240.47 \\ 
wordpress.com &          239.23 \\ 
apple.com &              236.77 \\ 
twitch.tv &              220.57 \\ 
huffpost.com &            219.8 \\ 
tumblr.com &             217.38 \\ 
birdops.com &            210.94 \\ 
snpy.tv &                 210.9 \\ 
foursquare.com &         207.91 \\ 
amazon.com &             206.87 \\ 
theguardian.com &        196.59 \\ 
bbc.co.uk &              195.18 \\ 
atmlb.com &              193.08 \\ 
gofundme.com &           181.99 \\ 
linkedin.com &           175.83 \\ 
uapp.ly &                171.29 \\ 
curiouscat.me &          169.83 \\ 
pinterest.com &          166.41 \\ 
buzzfeed.com &           165.26 \\ 
usa.gov &                164.36 \\ 
conta.cc&               160.76 \\ 
senate.gov &             159.12 \\ 
medium.com &             157.13 \\ 
eventbrite.com &          150.9 \\ 
ble.ac &                 143.32 \\ 
wsj.com &                143.09 \\ 
cnn.com &                139.38 \\ 
hudl.com &               133.94 \\ 
politico.com &            128.7 \\ 
usatoday.com &           126.92 \\ 
forbes.com &             126.03 \\ 
iacr.org &               122.46 \\ 
crowdfireapp.com &       122.11 \\ 
thehill.com &             121.6 \\ 
change.org &             121.29 \\ 
untappd.com &            121.19 \\ 
nfl.com &                117.88 \\ 
npr.org &                114.22 \\ 
abcnews.com &            114.22 \\ 
apple.news &             109.49 \\ 
bible.com &              109.38 \\  
        \end{tabular}
    \end{minipage} 
    \begin{minipage}{.33\linewidth}
      \centering
        \begin{tabular}{l|r}
            \textbf{censored} ($\gamma_{\textrm{max}}$=0.4)  & counts \\ \hline
\textcolor{red}{twitter.com}                        & 1571276 \\
\textcolor{red}{youtube.com}                         & 211489 \\
\textcolor{red}{facebook.com}                        & 106082 \\
\textcolor{red}{vine.co}                              & 25835 \\
\textcolor{red}{ift.tt}                               & 22096 \\
\textcolor{red}{google.com}                           & 17822 \\
\textcolor{red}{blogspot.com}                         & 16239 \\
\end{tabular}
    \end{minipage} 
\end{table}

\begin{table}[!htb]
    \caption*{\huge{\textbf{\underline{Russia (2018)}}}}
    \begin{minipage}{.33\linewidth}
      \centering
        \begin{tabular}{l|r}
             \textbf{indicator} & counts \\ \hline
UNCOMMON &                 101869 \\ 
instagram.com &              27361 \\ 
washingtonpost.com &         16305 \\ 
foxnews.com &                12065 \\ 
nytimes.com &                 6143 \\ 
usatoday.com &                4831 \\ 
huffpost.com &                4520 \\
\end{tabular}
    \end{minipage}%
    \begin{minipage}{.33\linewidth}
      \centering
        \begin{tabular}{l|r}
            \textbf{indicator}  & tf-idf \\ \hline
 instagram.com &          501.11 \\ 
cnn.com &                481.74 \\ 
wikipedia.org &          263.78 \\ 
oprf.ru &                257.69 \\ 
sdelanounas.ru &         216.52 \\ 
swarmapp.com &           180.14 \\ 
dni.ru &                 137.09 \\ 
nytimes.com &            118.84 \\ 
rvns.co &                108.87 \\ 
ukraina.ru &              99.21 \\ 
tumblr.com &               98.1 \\ 
iz.ru &                   84.76 \\ 
crowdfireapp.com &        57.37 \\ 
theguardian.com &         56.06 \\ 
wordpress.com &           54.68 \\ 
huffpost.com &            53.88 \\ 
whitehouse.gov &          48.24 \\ 
apple.com &               47.39 \\ 
change.org &              46.88 \\ 
snpy.tv &                 44.71 \\ 
soundcloud.com &          43.95 \\ 
washingtonpost.com &      42.22 \\ 
sky.com &                 40.49 \\ 
foxnews.com &             39.41 \\ 
afp.com &                 39.19 \\ 
bbc.co.uk &               39.09 \\ 
wsj.com &                 38.62 \\ 
golos-dnr.ru &            35.71 \\ 
amazon.com &              32.78 \\ 
usatoday.com &            29.25 \\ 
rusnext.ru &              27.76 \\ 
salon.com &               23.24 \\ 
espn.com &                22.94 \\ 
pscp.tv &                 22.73 \\ 
msn.com &                 22.29 \\ 
rawstory.com &            21.05 \\ 
nbcnews.to &              20.51 \\ 
vice.com &                18.93 \\ 
sumall.com &              18.38 \\ 
politico.com &            18.36 \\ 
buzzfeed.com &            18.19 \\ 
bloomberg.com &           18.17 \\ 
time.com &                17.77 \\ 
iheart.com &              17.55 \\ 
freebeacon.com &          17.09 \\ 
apnews.com &              16.67 \\ 
uapp.ly &                 16.66 \\ 
healthcare.gov &          16.47 \\ 
businessinsider.com &     15.61 \\ 
aka.ms &                  15.27 \\ 
economist.com &           15.03 \\ 
atmlb.com &               14.67 \\ 
        \end{tabular}
    \end{minipage} 
    \begin{minipage}{.33\linewidth}
      \centering
        \begin{tabular}{l|r}
            \textbf{censored} ($\gamma_{\textrm{max}}$=0.4)  & counts \\ \hline
\textcolor{red}{livejournal.com}       & 617933 \\
\textcolor{red}{riafan.ru}             & 452984 \\
\textcolor{red}{twitter.com}           & 285859 \\
\textcolor{red}{ria.ru}                & 166088 \\
\textcolor{red}{yandex.ru}             & 152580 \\
\textcolor{red}{gazeta.ru}             & 129518 \\
\textcolor{red}{youtube.com}           & 122115 \\
\end{tabular}
    \end{minipage} 
\end{table}

\begin{table}[!htb]
    \caption*{\huge{\textbf{\underline{China (2019)}}}}
    \begin{minipage}{.33\linewidth}
      \centering
        \begin{tabular}{l|r}
             \textbf{indicator} & counts \\ \hline
UNCOMMON &             131091 \\ 
instagram.com &          17516 \\ 
nytimes.com &             8756 \\ 
tumblr.com &              6757 \\ 
soundcloud.com &          5850 \\ 
bbc.co.uk &               3153 \\ 
uapp.ly &                 2514 \\ 
wordpress.com &           1902 \\ 
forbes.com &              1824 \\ 
etsy.com &                1720 \\ 
latimes.com &             1598 \\ 
pscp.tv &                 1486 \\ 
amazon.com &              1205 \\ 
huffpost.com &            1109 \\ 
espn.com &                1044 \\ 
twitch.tv &                990 \\ 
fllwrs.com &               906 \\ 
apple.com &                878 \\ 
wsj.com &                  848 \\ 
cnn.com &                  818 \\ 
crowdfireapp.com &         741 \\ 
foursquare.com &           666 \\ 
tcrn.ch &                  656 \\ 
foxnews.com &              617 \\ 
datpiff.com &              605 \\ 
yahoo.com &                569 \\ 
theguardian.com &          539 \\ 
engadget.com &             532 \\ 
curiouscat.me &            523 \\ 
gigam.es &                 492 \\ 
feedly.com &               398 \\ 
twittascope.com &          396 \\ 
washingtonpost.com &       383 \\ 
typepad.com &              379 \\ 
swarmapp.com &             345 \\ 
justunfollow.com &         338 \\ 
pinterest.com &            328 \\ 
pulse.me &                 327 \\ 
military.com &             297 \\ 
mashable.com &             290 \\ 
linkedin.com &             287 \\ 
flickr.com &               274 \\ 
nba.com &                  262 \\ 
patch.com &                257 \\ 
bloomberg.com &            246 \\ 
microsoft.com &            233 \\ 
natgeo.com &               226 \\ 
justgiving.com &           224 \\ 
usatoday.com &             224 \\ 
usa.gov &                  212 \\ 
foodandwine.com &          205 \\ 
thinkprogress.org &        197 \\      
        \end{tabular}
    \end{minipage}%
    \begin{minipage}{.33\linewidth}
      \centering
        \begin{tabular}{l|r}
            \textbf{indicator}  & tf-idf \\ \hline
instagram.com &          110.62 \\ 
forbes.com &             101.06 \\ 
wsj.com &                 83.31 \\ 
nba.com &                 74.34 \\ 
pscp.tv &                  57.7 \\ 
twitch.tv &               49.74 \\ 
tumblr.com &              45.29 \\ 
uapp.ly &                  43.4 \\ 
nytimes.com &             42.68 \\ 
curiouscat.me &           39.99 \\ 
time.com &                39.77 \\ 
soundcloud.com &          34.44 \\ 
natgeo.com &              34.39 \\ 
apple.com &               33.13 \\ 
amazon.com &              30.43 \\ 
app.link &                27.99 \\ 
scmp.com &                25.93 \\ 
reddit.com &              25.87 \\ 
wordpress.com &           23.63 \\ 
amazonaws.com &           23.25 \\ 
bbc.co.uk &               22.29 \\ 
fllwrs.com &              20.95 \\ 
foodandwine.com &         20.42 \\ 
crowdfireapp.com &        19.81 \\ 
variety.com &             18.58 \\ 
onlyfans.com &            18.12 \\ 
cnn.com &                 17.64 \\ 
yahoo.com &               17.48 \\ 
justunfollow.com &         17.4 \\ 
ew.com &                  16.95 \\ 
pinterest.com &           16.09 \\ 
politico.com &            15.68 \\ 
younow.com &              15.43 \\ 
flickr.com &              15.26 \\ 
linkfire.com &            14.85 \\ 
espn.com &                14.37 \\ 
fiverr.com &              14.09 \\ 
etsy.com &                13.74 \\ 
theguardian.com &         13.05 \\ 
cbsnews.com &             12.99 \\ 
shareaholic.com &         12.79 \\ 
cntraveler.com &          12.54 \\ 
foursquare.com &          12.32 \\ 
facebook.com &            12.06 \\ 
medium.com &              11.75 \\ 
ebay.com &                11.33 \\ 
nbcsports.com &           10.98 \\ 
eonline.com &             10.71 \\ 
bleacherreport.com &       10.7 \\ 
pagesix.com &             10.48 \\ 
theglobeandmail.com &     10.46 \\ 
washingtonpost.com &      10.44 \\              
        \end{tabular}
    \end{minipage}
        \begin{minipage}{.33\linewidth}
      \centering
        \begin{tabular}{l|r}
            \textbf{censored} ($\gamma_{\textrm{max}}$=0.4)  & counts \\ \hline
\textcolor{red}{twitter.com}                     & 526189 \\
\textcolor{red}{dld.bz}                          & 195629 \\
\textcolor{red}{youtube.com}                     & 137703 \\
\textcolor{red}{feedburner.com}                  & 115155 \\
\textcolor{red}{miss50percent.de}                & 107453 \\
\textcolor{red}{happymuslimfamily.net}            & 62821 \\
\textcolor{red}{ift.tt}                           & 52636 \\
\end{tabular}
    \end{minipage} 
\end{table}

\begin{table}[!htb]
    \caption*{\huge{\textbf{\underline{Iran (2019)}}}}
    \begin{minipage}{.33\linewidth}
      \centering
        \begin{tabular}{l|r}
             \textbf{indicator} & counts \\ \hline
UNCOMMON &                 511760 \\ 
instagram.com &              13943 \\ 
theguardian.com &             2506 \\ 
amazon.com &                  1980 \\ 
nytimes.com &                 1977 \\ 
wordpress.com &               1768 \\ 
huffpost.com &                1602 \\
\end{tabular}
    \end{minipage}%
    \begin{minipage}{.33\linewidth}
      \centering
        \begin{tabular}{l|r}
            \textbf{indicator}  & tf-idf \\ \hline
 instagram.com &             242.26 \\ 
theguardian.com &           184.91 \\ 
pscp.tv &                   158.72 \\ 
amazon.com &                 97.03 \\ 
mirror.co.uk &               95.83 \\ 
reddit.com &                 92.89 \\ 
soundcloud.com &             87.84 \\
\end{tabular}
    \end{minipage} 
        \begin{minipage}{.33\linewidth}
      \centering
        \begin{tabular}{l|r}
            \textbf{censored} ($\gamma_{\textrm{max}}$=0.4)  & counts \\ \hline
\textcolor{red}{awdnews.com}                                         & 332374 \\
\textcolor{red}{twitter.com}                                         & 241386 \\
\textcolor{red}{irib.ir}                                             & 217793 \\
\textcolor{red}{parstoday.com}                                       & 153488 \\
\textcolor{red}{tel-avivtimes.com}                                   & 143601 \\
\textcolor{red}{countdown2040.com}                                   & 139262 \\
\textcolor{red}{nilenetonline.com}                                   & 113892 \\
\end{tabular}
    \end{minipage} 
\end{table}

\begin{table}[!htb]
    \caption*{\huge{\textbf{\underline{Russia (2020)}}}}
    \begin{minipage}{.33\linewidth}
      \centering
        \begin{tabular}{l|r}
             \textbf{indicator} & counts \\ \hline
UNCOMMON &             33522 \\ 
instagram.com &         22190 \\ 
foursquare.com &         7586 \\ 
swarmapp.com &           2104 \\ 
furnishhome.ru &         1151 \\ 
iz.ru &                   615 \\ 
dni.ru &                  552 \\ 
ft.com &                  532 \\ 
crowdfireapp.com &        521 \\ 
tumblr.com &              518 \\ 
nytimes.com &             467 \\ 
wsj.com &                 405 \\ 
sdelanounas.ru &          373 \\ 
wordpress.com &           365 \\ 
change.org &              329 \\ 
sumall.com &              252 \\ 
theguardian.com &         236 \\ 
linkedin.com &            227 \\ 
apple.com &               206 \\ 
medium.com &              203 \\ 
washingtonpost.com &      197 \\ 
wikipedia.org &           181 \\ 
csmonitor.com &           173 \\ 
izhgpk.ru &               169 \\ 
soundcloud.com &          164 \\ 
forbes.com &              155 \\ 
fllwrs.com &              149 \\ 
newsomsk.ru &             134 \\ 
huffpost.com &            122 \\ 
gizmodo.com &             101 \\ 
thenextweb.com &           88 \\ 
nike.com &                 87 \\ 
awe.sm &                   74 \\ 
bloomberg.com &            72 \\ 
bbc.co.uk &                71 \\ 
pscp.tv &                  71 \\ 
cnn.com &                  56 \\ 
thinkprogress.org &        54 \\ 
ustream.tv &               53 \\ 
anekdot.ru &               52 \\ 
time.com &                 51 \\ 
diletant.media &           45 \\ 
thediplomat.com &          42 \\ 
foreignpolicy.com &        41 \\ 
motherjones.com &          40 \\ 
whitehouse.gov &           39 \\ 
valuemytweets.com &        36 \\ 
oprf.ru &                  36 \\ 
buzzfeed.com &             35 \\ 
theverge.com &             35 \\ 
quora.com &                34 \\ 
thehill.com &              33 \\ 

        \end{tabular}
    \end{minipage}%
    \begin{minipage}{.33\linewidth}
      \centering
        \begin{tabular}{l|r}
            \textbf{indicator}  & tf-idf \\ \hline
 iz.ru &                   293.69 \\ 
dni.ru &                  193.44 \\ 
instagram.com &            144.6 \\ 
quora.com &               126.54 \\ 
thenextweb.com &          115.03 \\ 
furnishhome.ru &           107.5 \\ 
foursquare.com &            78.0 \\
\end{tabular}
    \end{minipage} 
        \begin{minipage}{.33\linewidth}
      \centering
        \begin{tabular}{l|r}
            \textbf{censored} ($\gamma_{\textrm{max}}$=0.4)  & counts \\ \hline
\textcolor{red}{lenta.ru}              & 213528 \\
\textcolor{red}{yandex.ru}             & 176599 \\
\textcolor{red}{rambler.ru}             & 72433 \\
\textcolor{red}{newkaliningrad.ru}      & 62448 \\
\textcolor{red}{161.ru}                 & 42756 \\
\textcolor{red}{profile.ru}             & 41755 \\
\textcolor{red}{ufa1.ru}                & 39093 \\
\end{tabular}
    \end{minipage} 
\end{table}

\begin{table}[!htb]
    \caption*{\huge{\textbf{\underline{China (2020)}}}}
    \begin{minipage}{.33\linewidth}
      \centering
        \begin{tabular}{l|r}
             \textbf{indicator} & counts \\ \hline
UNCOMMON &             1693 \\ 
axios.com &              231 \\ 
medium.com &              36 \\ 
pscp.tv &                 31 \\ 
bloomberg.com &           30 \\ 
cnn.com &                 26 \\ 
nytimes.com &             24 \\ 
amazon.com &              21 \\ 
foxnews.com &             18 \\ 
linkfire.com &            17 \\ 
instagram.com &           13 \\ 
theguardian.com &         10 \\ 
washingtonpost.com &      10 \\ 
cbsnews.com &             10 \\ 
weebly.com &               9 \\ 
wordpress.com &            9 \\ 
nu.nl &                    9 \\ 
time.com &                 9 \\ 
usatoday.com &             9 \\ 
lasvegassun.com &          7 \\ 
yahoo.com &                7 \\ 
wsj.com &                  6 \\ 
apple.com &                5 \\ 
economist.com &            5 \\ 
app.link &                 5 \\ 
bbc.co.uk &                5 \\ 
swarmapp.com &             4 \\ 
change.org &               4 \\ 
katu.com &                 4 \\ 
rfa.org &                  4 \\ 
broadway.com &             4 \\ 
natgeo.com &               3 \\ 
tumblr.com &               3 \\ 
adobe.com &                3 \\ 
quora.com &                3 \\ 
google.com &               3 \\ 
insider.com &              3 \\ 
soundcloud.com &           3 \\ 
variety.com &              3 \\ 
nikkei.com &               2 \\ 
diletant.media &           2 \\ 
wistia.com &               2 \\ 
obama.org &                2 \\ 
mediamatters.org &         2 \\ 
meetedgar.com &            2 \\ 
marvel.com &               2 \\ 
sns.mx &                   2 \\ 
people.com &               2 \\ 
petapixel.com &            2 \\ 
eonline.com &              2 \\ 
forbes.com &               2 \\ 
fiverr.com &               2 \\ 
  
        \end{tabular}
    \end{minipage}%
    \begin{minipage}{.33\linewidth}
      \centering
        \begin{tabular}{l|r}
            \textbf{indicator}  & tf-idf \\ \hline
 axios.com &             164.41 \\ 
linkfire.com &           29.84 \\ 
adobe.com &              14.58 \\ 
bloomberg.com &          13.46 \\ 
wordpress.com &          10.48 \\ 
weebly.com &             10.32 \\ 
medium.com &             10.03 \\ 
amazon.com &              9.86 \\ 
pscp.tv &                 9.31 \\ 
people.com &              6.79 \\ 
fiverr.com &              6.69 \\ 
onlyfans.com &             6.6 \\ 
diletant.media &           6.2 \\ 
change.org &              5.69 \\ 
app.link &                5.51 \\ 
apple.com &               5.43 \\ 
yahoo.com &               5.35 \\ 
nike.com &                5.27 \\ 
nytimes.com &              5.2 \\ 
wsj.com &                 4.07 \\ 
bbc.co.uk &               4.06 \\ 
cc.com &                  4.04 \\ 
washingtonpost.com &      3.96 \\ 
standard.co.uk &          3.81 \\ 
swarmapp.com &            3.67 \\ 
buzzfeednews.com &        3.66 \\ 
viewbug.com &             3.42 \\ 
fanlink.to &              3.27 \\ 
time.com &                3.08 \\ 
instagram.com &           3.03 \\ 
anchor.fm &               2.91 \\ 
tumblr.com &              2.88 \\ 
natgeo.com &              2.86 \\ 
theguardian.com &         2.36 \\ 
uscis.gov &               2.35 \\ 
pandora.com &             2.28 \\ 
cnn.com &                  2.1 \\ 
justice.gov &             2.06 \\ 
foxnews.com &             2.03 \\ 
cnet.com &                1.94 \\ 
lasvegassun.com &         1.82 \\ 
soundcloud.com &          1.81 \\ 
variety.com &             1.62 \\ 
pbs.org &                 1.47 \\ 
eonline.com &             1.32 \\ 
foreignpolicy.com &       1.25 \\ 
cbsnews.com &              1.2 \\ 
apnews.com &              1.12 \\ 
quora.com &               1.02 \\ 
usatoday.com &            0.81 \\ 
economist.com &           0.59 \\ 
etsy.com &                0.49 \\ 

        \end{tabular}
    \end{minipage} 
    \begin{minipage}{.33\linewidth}
      \centering
        \begin{tabular}{l|r}
            \textbf{censored} ($\gamma_{\textrm{max}}$=0.4)  & counts \\ \hline
\textcolor{red}{youtube.com}               & 8148 \\
\textcolor{red}{twitter.com}               & 1130 \\
\textcolor{red}{creaders.net}               & 829 \\
\textcolor{red}{temaretik.com}              & 669 \\
\textcolor{red}{rapradar.com}               & 503 \\
\textcolor{red}{pansci.asia}                & 348 \\
\textcolor{red}{dwnews.com}                 & 263 \\
\end{tabular}
    \end{minipage} 
    
\end{table}

\begin{table}[!htb]
    \caption*{\huge{\textbf{\underline{Iran (2021)}}}}
    \begin{minipage}{.33\linewidth}
      \centering
        \begin{tabular}{l|r}
             \textbf{indicator} & counts \\ \hline
UNCOMMON &                 1338 \\ 
nytimes.com &                373 \\ 
cbsnews.com &                213 \\ 
theguardian.com &            208 \\ 
wsj.com &                    203 \\ 
washingtonpost.com &         186 \\ 
cnn.com &                    186 \\
\end{tabular}
    \end{minipage}%
    \begin{minipage}{.33\linewidth}
      \centering
        \begin{tabular}{l|r}
            \textbf{indicator}  & tf-idf \\ \hline
 nytimes.com &              19.49 \\ 
wsj.com &                  18.16 \\ 
instagram.com &            10.53 \\ 
theguardian.com &           9.81 \\ 
rawstory.com &               9.2 \\ 
whitehouse.gov &            7.52 \\ 
pscp.tv &                   7.21 \\
\end{tabular}
    \end{minipage} 
    \begin{minipage}{.33\linewidth}
      \centering
        \begin{tabular}{l|r}
            \textbf{censored} ($\gamma_{\textrm{max}}$=0.4)  & counts \\ \hline
\textcolor{red}{htv.mx}                    & 98259 \\
\textcolor{red}{hispantv.com}              & 43193 \\
\textcolor{red}{youtube.com}               & 35874 \\
\textcolor{red}{twitter.com}                & 4918 \\
\textcolor{red}{hispantv.ir}                & 1312 \\
\textcolor{red}{bing.com}                    & 451 \\
\textcolor{red}{facebook.com}                & 430 \\
\end{tabular}
    \end{minipage} 
\end{table}

\end{document}